%% file: main.tex
\documentclass[11pt]{article}

\usepackage{hydrosophy}
\usepackage{xurl}
\usepackage{multirow}
\usepackage{tabularx}
\usepackage{pifont}
\usepackage{marvosym}
\usepackage{accsupp}
\usepackage{tikz}

\newcommand{\corrmark}{\BeginAccSupp{method=hex,unicode,ActualText=2709}\Letter\EndAccSupp{}}
\newcommand{\cmark}{\textcolor{green!60!black}{\ding{51}}}
\newcommand{\xmark}{\textcolor{red!75!black}{\ding{55}}}
\newcommand{\pmark}{\textcolor{orange!85!black}{$\triangle$}}

\definecolor{bkAct}{HTML}{258D81}
\definecolor{bkKnow}{HTML}{3475B7}
\definecolor{bkExp}{HTML}{8057A2}
\definecolor{bkReas}{HTML}{DD7839}
\definecolor{bkInk}{HTML}{24343A}
\definecolor{bkMuted}{HTML}{5D7076}
\definecolor{bkRule}{HTML}{A5B2B5}
\definecolor{bkHead}{HTML}{EAF0F1}
\definecolor{bkZebra}{HTML}{F5F8F8}
\definecolor{bkGap}{HTML}{B9534C}
\colorlet{bkImg}{bkAct}
\colorlet{bkVid}{bkExp}
\colorlet{bkActLight}{bkAct!55}
\colorlet{bkCatChem}{bkReas}
\colorlet{bkCatSci}{bkExp}
\colorlet{bkCatEmb}{bkAct}
\colorlet{bkCatGen}{bkKnow}
\DeclareRobustCommand{\bkdot}[1]{\tikz[baseline=-0.55ex]\fill[#1] (0,0) circle (0.26em);}
\newcommand{\bkbar}[4][1.2cm]{\tikz[baseline=-0.55ex]{%
  \fill[#4!15, rounded corners=0.5pt] (0,-0.33em) rectangle (#1,0.33em);%
  \fill[#4, rounded corners=0.5pt] (0,-0.33em) rectangle ({#2/#3*#1},0.33em);}}
\newcommand{\bktoprule}{\arrayrulecolor{bkInk}\toprule\arrayrulecolor{bkRule}}
\newcommand{\bkbottomrule}{\arrayrulecolor{bkInk}\bottomrule\arrayrulecolor{black}}
\newcommand{\bkhead}{\rowcolor{bkHead}}
\newcommand{\bkgroup}[3]{\rowcolor{#1!9}\multicolumn{#2}{l}{\textcolor{#1!80!black}{\textbf{#3}}}}
\newcommand{\bkmuted}[1]{\textcolor{bkMuted}{#1}}
\newcommand{\bksetup}[1]{\setlength{\aboverulesep}{0pt}\setlength{\belowrulesep}{0pt}\renewcommand{\arraystretch}{#1}}
\DeclareRobustCommand{\bkpill}[2][bkMuted]{\tikz[baseline=(p.base)]\node[inner xsep=2.4pt, inner ysep=1pt, text height=1.55ex, text depth=0.4ex, rounded corners=2.2pt, fill=#1!13, text=#1!80!black, font=\fontsize{7.6}{9}\selectfont] (p) {#2};}
\DeclareRobustCommand{\bknum}[2]{\tikz[baseline=(n.base)]\node[circle, fill=#1, text=white, font=\scriptsize\bfseries, inner sep=0pt, minimum size=1.3em] (n) {#2};}
\newcommand{\bkciaxis}{\tikz[baseline=0pt, x=0.082cm, y=1pt]{%
  \useasboundingbox (0,-1) rectangle (37,6);%
  \foreach \g in {30,40,50,60}{\node[font=\scriptsize, text=bkMuted, inner sep=0pt, anchor=base] at (\g-25,0) {\g};}}}
\newcommand{\bkci}[4]{\tikz[baseline=0pt, x=0.082cm, y=1pt]{%
  \useasboundingbox (0,0) rectangle (37,6);%
  \foreach \g in {30,40,50,60}{\draw[bkRule!50, line width=0.3pt] (\g-25,-3.6) -- (\g-25,8.2);}%
  \draw[#4, line width=1.2pt, line cap=round] (#1-25,2.3) -- (#2-25,2.3);%
  \fill[white] (#3-25,2.3) circle (2.1pt);%
  \fill[#4!80!black] (#3-25,2.3) circle (1.5pt);}}
\newcommand{\bkdevaxis}{\tikz[baseline=0pt, x=0.62cm, y=1pt]{%
  \useasboundingbox (-3.5,-1) rectangle (3.5,6);%
  \foreach \v/\l in {-3/$-3$,-2/$-2$,-1/$-1$,0/$0$,1/$+1$,2/$+2$,3/$+3$}{\node[font=\scriptsize, text=bkMuted, inner sep=0pt, anchor=base] at (\v,0) {\l};}}}
\newcommand{\bkdev}[3]{\tikz[baseline=0pt, x=0.62cm, y=1pt]{%
  \useasboundingbox (-3.5,0) rectangle (3.5,6);%
  \fill[#3!14] ({-2*#1},-0.8) rectangle ({2*#1},5.4);%
  \fill[#3!34] ({-#1},-0.8) rectangle ({#1},5.4);%
  \draw[bkMuted!70, line width=0.35pt] (0,-3.9) -- (0,8.9);%
  \fill[white] (#2,2.3) circle (2.2pt);%
  \fill[#3!85!black] (#2,2.3) circle (1.6pt);}}
\newcommand{\bkph}[1]{\textcolor{bkReas!85!black}{\{#1\}}}
\newcommand{\bkstack}[2]{\begin{tabular}[b]{@{}c@{}}#1\\#2\end{tabular}}
\newtcolorbox{bkprompt}[3][bkImg]{enhanced, colback=#1!3, colframe=#1!75!bkInk,
  boxrule=0.5pt, arc=2pt, colbacktitle=#1!80!bkInk, coltitle=white,
  fonttitle=\small\bfseries, title={#2\hfill\mdseries\itshape #3},
  toptitle=1.5pt, bottomtitle=1.5pt, lefttitle=7pt, righttitle=7pt,
  fontupper=\small\ttfamily\raggedright, left=7pt, right=7pt, top=4pt, bottom=4pt,
  before skip=9pt, after skip=9pt}

\newcommand{\bkgithub}{\raisebox{-1.5pt}{\includegraphics[height=1.05em]{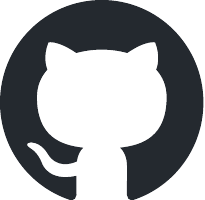}}}
\newcommand{\bkhf}{\raisebox{-1.5pt}{\includegraphics[height=1.05em]{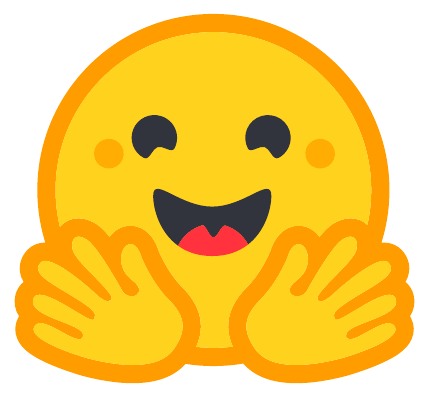}}}

\title{BEAKER: An Expert-Curated Benchmark for Embodied Brains in Self-Driving Chemical Laboratories}
\shorttitle{BEAKER}
\author[1,2,3,*,\textdagger]{Fei Lin}
\author[1,2,*]{Tengchao Zhang}
\author[1,3,*]{Ziyang Gong}
\author[4]{Xiaotong Yu}
\author[2]{Bohan Zhang}
\author[3]{Yifan Zhou}
\author[3]{Qihao Yang}
\author[5]{Ji Dai}
\author[2]{Dong Li}
\author[2]{Yining Jiang}
\author[2]{Qinghua Ni}
\author[2]{Jun Huang}
\author[6]{Qiang Zhang}
\author[7]{Yue Zhou}
\author[3]{Zhihang Zhong}
\author[8,\corrmark]{Yonglin Tian}
\author[2,8,\corrmark]{Fei-Yue Wang}
\author[3,\corrmark]{Xue Yang}

\affiliation[1]{HydroSophy}
\affiliation[2]{Macau University of Science and Technology}
\affiliation[3]{Shanghai Jiao Tong University}
\affiliation[4]{Institute of Process Engineering, Chinese Academy of Sciences}
\affiliation[5]{Ningbo University}
\affiliation[6]{Dalian University of Technology}
\affiliation[7]{East China Normal University}
\affiliation[8]{Institute of Automation, Chinese Academy of Sciences}

\authornote[*]{Equal contribution.}
\authornote[\textdagger]{Project lead.}
\authornote[\corrmark]{Corresponding authors.}

\metadata[\bkgithub\ Code]{\url{https://github.com/HydroSophy/BEAKER}}
\metadata[\bkhf\ Dataset]{\url{https://huggingface.co/datasets/HydroSophyTech/BEAKER}}

\begin{document}

\maketitle

\input{sections/0_abstract}

\input{sections/1_introduction}
\input{sections/2_related_work}
\input{sections/3_BEAKER}
\input{sections/4_experiments}
\input{sections/5_failure_attribution}
\input{sections/6_limitations}
\input{sections/7_conclusion}

\bibliographystyle{unsrtnat}
\bibliography{references}

\hsappendix
\input{sections/appendix}

\end{document}

%% file: sections/0_abstract.tex
\begin{abstract}
Self-Driving Chemical Laboratories (SDCLs) are moving chemical experimentation toward embodied automation, where machines must perceive laboratory scenes, reason about experimental states, and act under physical, procedural, and safety constraints. Although Multimodal Large Language Models (MLLMs) are increasingly considered as embodied brains for such systems, existing laboratory benchmarks mainly focus on safety, anomaly detection, or general scientific understanding, leaving their execution-oriented embodied capabilities underexplored. To fill this gap, we introduce \textbf{BEAKER}, a \textbf{B}enchmark for evaluating \textbf{E}mbodied \textbf{A}ctionability, \textbf{K}nowledge, \textbf{E}xperimentation, and \textbf{R}easoning in SDCLs. BEAKER contains 1,000 expert-curated question-answer samples across 11 image and video task types, covering key judgments required for chemical laboratory execution, including affordance grounding, trajectory planning, equipment, state, and safety recognition, apparatus setup, operation and step reasoning, and temporal procedure reasoning. All questions and annotations are manually created and reviewed by chemistry and embodied Artificial Intelligence (AI) experts, without AI-assisted generation or annotation. We evaluate 28 general-purpose, science-oriented, chemistry-specialized, and embodied MLLMs in an offline, single-turn setting. Results show that current models can follow coarse experimental procedures in videos, but still struggle to ground actions into operable regions and feasible trajectories, to recognize fine-grained laboratory apparatus, states, and safety norms, and to order steps in mid-length procedures. BEAKER thus reveals a clear gap between coarse procedural understanding and the precise, action-grounded perception required for laboratory execution, providing a diagnostic benchmark for future MLLMs in SDCLs.
\end{abstract}

%% file: sections/1_introduction.tex
\section{Introduction}

\input{figures/teaser}

Self-Driving Chemical Laboratories (SDCLs) are transforming chemical experimentation from manual trial-and-error into closed-loop autonomous discovery, where automated devices execute experiments, observe outcomes, and use data-driven feedback to guide subsequent operations~\cite{tom2024self,pilon2026flexible,burger2020mobile,dai2024autonomous,abolhasani2023rise,leonov2024selfoptimizing,canty2025science,leong2025steering,lin2025autonomous}. Such systems require more than predefined protocol execution: automated devices must perceive laboratory scenes, understand experimental states, and make execution-relevant judgments under physical, procedural, and safety constraints. Recent advances in Multimodal Large Language Models (MLLMs) serving as embodied brains provide a promising paradigm for this goal, aiming to connect multimodal perception with affordance reasoning, task progress, physical constraints, and action-oriented decision-making~\cite{robobrain2025robobrain,gong2026ace,team2026ace,vebrain2025vebrain,yang2026visual,robobrain2025robobrain2,yuan2026embodied,wang2025affordbot}. However, existing embodied brain studies mainly focus on general robotic and embodied scenarios, while scientific experimental automation remains much less explored.

This requirement is non-trivial because chemical experiments are not static visual scenes but procedural, operable, and safety-constrained embodied processes, in which visual information becomes meaningful only in the context of the current execution. The same apparatus, liquid level, connection state, or material change may imply different actions, risks, or experimental outcomes depending on the current step, operability, connection correctness, and safety requirements. Recognizing a flask, a liquid interface, or a connected tube is therefore insufficient if a model cannot determine whether the apparatus is ready to operate, whether the connection supports the next step, or whether an observed state change indicates normal progress or potential failure. This raises a central question: can current MLLMs move from laboratory scene understanding to execution-grounded reasoning in chemical~experiments?

Recent studies have begun to explore MLLMs in SDCLs, demonstrating their potential in robotic chemical operations, experimental state monitoring, and safety risk identification~\cite{zhou2025localization,zhang2025robochemist,veeramani2025prevent,wei2026agentchemist,darvish2024organa,song2025multiagent,panapitiya2025autolabs,ruan2024endtoend}. Meanwhile, emerging laboratory-oriented multimodal benchmarks mainly focus on anomaly detection, hazard identification, safety planning, or general scientific understanding~\cite{lin2025visual, lin2025vlm, sun2026labshield, alampara2025probing, xu2025expvid}. While valuable, they do not systematically assess the embodied capabilities required for automated experimental execution. A suitable benchmark for SDCL-oriented embodied brains should evaluate whether MLLMs can identify operable regions, understand laboratory apparatus and states, track execution steps, and reason about temporal changes and experimental~outcomes.

To fill this gap, we introduce \textbf{BEAKER}, short for \textbf{B}enchmark for \textbf{E}mbodied \textbf{A}ctionability, \textbf{K}nowledge, \textbf{E}xperimentation, and \textbf{R}easoning in SDCLs (Fig.~\ref{fig:teaser}). BEAKER decomposes embodied chemical laboratory execution into four core capabilities, each evaluated through expert-designed Question-Answering (QA) tasks. \emph{Actionability} concerns whether a model understands operable regions, interaction possibilities, and motion feasibility. \emph{Knowledge} concerns whether it recognizes laboratory apparatus, experimental states, and procedural norms. \emph{Experimentation} concerns whether it understands experimental operations, apparatus connections, and step progression. \emph{Reasoning} concerns whether it can infer temporal changes, procedural consequences, and experimental outcomes. Together, these dimensions provide an execution-oriented framework for evaluating whether MLLMs can support automated chemical experimentation beyond passive scene understanding.

Specifically, we collect more than 10,000 publicly available real-world images and videos related to chemical laboratories. Most samples are collected from first-person or experiment-execution-related viewpoints, covering diverse chemical laboratory scenes and workbenches. All samples are jointly selected, reviewed, and annotated by chemistry experts and embodied Artificial Intelligence (AI) experts. The resulting BEAKER dataset contains 1,000 carefully designed QA samples, covering seven image-based task types and four video-based task types. All questions are manually authored by human experts according to chemical laboratory execution contexts, without using AI tools for sample selection, question generation, or annotation, thereby reducing the self-bias that may arise from the MLLM-as-benchmark-generator paradigm. These tasks do not merely test chemical commonsense or laboratory object recognition; instead, they evaluate, in an offline, single-turn setting, the execution-grounded perception and reasoning on which closed-loop SDCL execution depends.

We further evaluate 28 mainstream MLLMs on BEAKER, including general-purpose, science-oriented, chemistry-specialized, and embodied multimodal models. The results reveal that current models can follow coarse procedural progress in experimental videos, but still struggle with execution-oriented judgments. In particular, localizing fine-grained operable regions, planning feasible trajectories, recognizing domain-specific apparatus, reagent states, and safety norms, and ordering steps in mid-length procedures remain challenging, while science-oriented and chemistry-specialized models do not consistently outperform general-purpose MLLMs. These findings reveal that domain knowledge alone is insufficient for SDCL execution and that current models still need to ground laboratory observations into operable actions, procedural progress, and experimental outcomes.

In summary, our contributions are as follows:

\begin{itemize}[leftmargin=*]
    \item We introduce \textbf{BEAKER}, an offline, single-turn benchmark for evaluating multimodal embodied brains in SDCLs, which organizes chemical laboratory automation capabilities into four core dimensions: \emph{Actionability}, \emph{Knowledge}, \emph{Experimentation}, and \emph{Reasoning}.
    \item We construct a multimodal dataset with 1,000 QA samples across 11 key tasks. All samples are manually selected and annotated by domain experts to ensure that the evaluation remains grounded in realistic chemical laboratory execution contexts.
    \item We evaluate 28 mainstream MLLMs on BEAKER and reveal a clear execution gap: current models can follow coarse experimental procedures, but still struggle with fine-grained action grounding, domain-specific laboratory perception, and mid-length procedural reasoning for SDCL automation.
\end{itemize}

%% file: figures/teaser.tex
\begin{figure*}[t]
    \centering
    \includegraphics[width=1.0\textwidth]{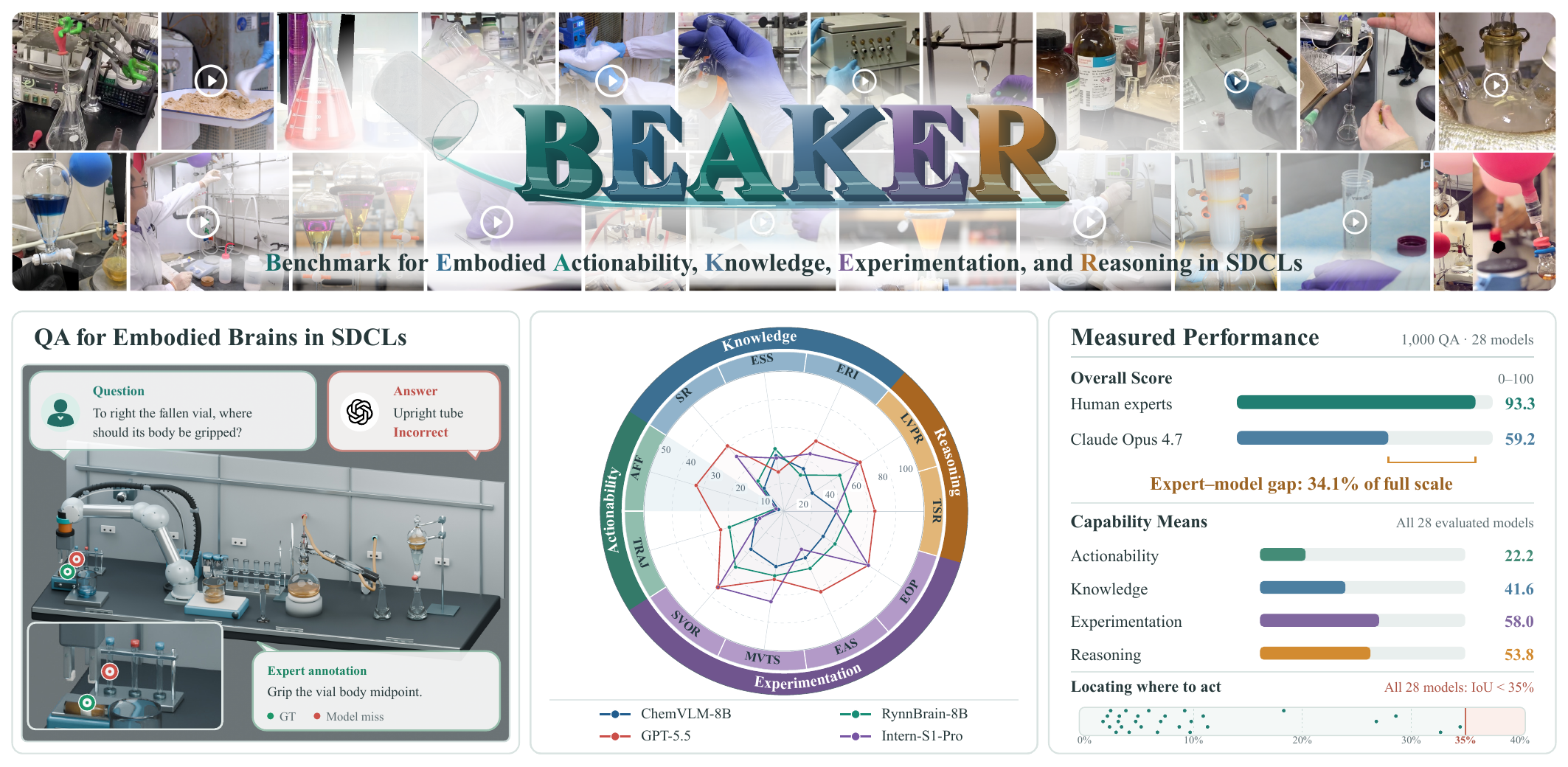}
    \caption{\textbf{Overview of BEAKER and key evaluation findings.} The radar chart compares four representative models across 11 tasks grouped by capability. All axes use a 0--100 scale except AFF, which uses a 0--50 scale due to uniformly low scores across all models. AFF: Affordance Grounding; TRAJ: Trajectory Planning; ERI: Equipment and Reagent Identification; ESS: Experimental Safety and Standardization; SR: State Recognition; EAS: Experimental Apparatus Setup; EOP: Experimental Operations; SVOR: Short-Video Operation Reasoning; MVTS: Mid-Video Task-Step Reasoning; TSR: Temporal Sequence Reasoning; LVPR: Long-Video Experimental Procedure Reasoning; \mbox{GT: Ground Truth.}}
    \label{fig:teaser}
\end{figure*}

%% file: sections/2_related_work.tex
\section{Related Work}

\subsection{MLLM-enabled SDCL}

General-purpose MLLMs have shown strong abilities in image understanding, video reasoning, and long-context modeling~\cite{openai2023gpt4,anthropic2024claude3,gemini2024gemini15,kimi2025kimivl,qwen2025qwen25vl,wang2025internvl35,qian2026loopvl}. Science-oriented MLLMs further incorporate domain-specific scientific data and reasoning tasks~\cite{bai2025interns1,zou2026interns1pro}, while chemistry-specialized MLLMs target molecular structures, reaction images, and cross-modal chemical knowledge~\cite{li2025chemvlm,tan2025chemmllm,zhao2024chemdfmx,lin2026breaking}. In parallel, embodied MLLMs extend vision-language reasoning to physical-action scenarios, such as spatial understanding, affordance detection, trajectory prediction, and task planning~\cite{robobrain2025robobrain2,dang2026rynnbrain,mimo2025embodied,zhang2025pelicanvl}. These advances make MLLMs potential embodied brains for SDCLs. Recent works have applied MLLMs to robotic chemical operations~\cite{zhang2025robochemist,song2026chemvlmate}, execution-state monitoring with visual inspection and error correction~\cite{zhou2025localization}, and human-aware laboratory coordination~\cite{veeramani2026human}. However, real chemical experiments are dynamic, transparent, step-dependent, and safety-constrained, making visual grounding, state judgment, and error recovery still~challenging.

\subsection{Multimodal Embodied Brains and QA Benchmarks}

Multimodal embodied brains~\cite{robobrain2025robobrain,robobrain2025robobrain2,robobrain2026robobrain25,zhang2025pelicanvl,mimo2025embodied,vlaser2025vlaser,vebrain2025vebrain,nvidia2025cosmos,rovi2025rovi,robospatial2025robospatial,roboafford2025roboafford,yuan2026embodied,qu2025eo} are high-level cognitive models for physical-action scenarios, integrating vision, language, and task states for environment understanding, task planning, and execution feedback~\cite{tan2025roboos,ma2026survey,yang2026skillopt}. Multimodal QA benchmarks evaluate perception, knowledge, and reasoning from images, videos, or other visual inputs~\cite{antol2015vqa,hudson2019gqa,lu2022learn,yue2024mmmu,luo2026robobench}. Recent scientific benchmarks further cover chemical and materials experiment understanding, experiment video reasoning, \mbox{laboratory} embodied interaction, and robotic laboratory simulation~\cite{alampara2025probing,xu2025expvid,li2025labutopia,lan2025autobio,li2024chemistry3d}. However, benchmarks for the execution stage of SDCL-automated devices remain limited, as existing studies mainly address execution-process anomaly detection~\cite{lin2025visual,lin2025vlm} or laboratory safety reasoning~\cite{sun2026labshield}. As shown in Table~\ref{tab:benchmark_comparison}, BEAKER instead targets visual contexts in real chemical experiments and jointly evaluates Actionability, Knowledge, Experimentation, and Reasoning, assessing whether multimodal embodied brains can support automated SDCL execution.

\input{table/benchmark_comparison}

%% file: table/benchmark_comparison.tex
\begin{table*}[t]
\centering
\caption{\textbf{Comparison with related scientific and laboratory benchmarks.} Scale denotes the primary evaluation unit of each work: QA for question-answering samples, tasks for embodied or robotic tasks, pairs for image-text pairs, imgs for images, and Sim.\ env.\ for simulation environments. \cmark, \xmark, and \pmark\ denote full, no, and partial coverage, respectively. Human indicates whether all questions and annotations are authored by human experts without LLM/AI-assisted generation or annotation.}
\label{tab:benchmark_comparison}
\normalsize
\resizebox{\textwidth}{!}{
\begin{tabular}{lcccccccccc}
\toprule
\textbf{Benchmark}
& \textbf{Scale}
& \textbf{SDCL}
& \textbf{Real}
& \textbf{Image}
& \textbf{Video}
& \textbf{Action.}
& \textbf{Know.}
& \textbf{Exper.}
& \textbf{Reason.}
& \textbf{Human} \\
\midrule
MaCBench~\cite{alampara2025probing}
& 1.1k+ QA & \pmark & \cmark & \cmark & \xmark & \xmark & \cmark & \pmark & \cmark & \cmark \\
ExpVid~\cite{xu2025expvid}
& 7.8k QA & \pmark & \cmark & \xmark & \cmark & \xmark & \cmark & \cmark & \cmark & \pmark \\
LabUtopia~\cite{li2025labutopia}
& 30 tasks & \pmark & \xmark & \cmark & \xmark & \pmark & \cmark & \cmark & \cmark & \cmark \\
AutoBio~\cite{lan2025autobio}
& 16 tasks & \xmark & \xmark & \cmark & \xmark & \pmark & \cmark & \cmark & \pmark & \cmark \\
Chemistry3D~\cite{li2024chemistry3d}
& Sim. env. & \pmark & \xmark & \cmark & \xmark & \pmark & \cmark & \cmark & \pmark & \cmark \\
SDCL Visual Anomaly Dataset~\cite{lin2025visual}
& 2.8k pairs & \cmark & \cmark & \cmark & \xmark & \xmark & \pmark & \cmark & \cmark & \cmark \\
VLM-based Anomaly Detection~\cite{lin2025vlm}
& 1.0k imgs & \cmark & \cmark & \cmark & \xmark & \xmark & \pmark & \cmark & \cmark & \cmark \\
\textsc{LabShield}~\cite{sun2026labshield}
& 1.4k QA & \cmark & \cmark & \cmark & \xmark & \xmark & \cmark & \cmark & \cmark & \xmark \\
\midrule
\textbf{BEAKER (Ours)}
& 1.0k QA & \cmark & \cmark & \cmark & \cmark & \cmark & \cmark & \cmark & \cmark & \cmark \\
\bottomrule
\end{tabular}
}
\end{table*}

%% file: sections/3_BEAKER.tex
\section{BEAKER}
\label{sec:BEAKER}

\subsection{Design Principles}
\label{sec:principles}

Autonomous execution in SDCLs requires an embodied brain to close the loop from visual perception to physical action within chemically grounded contexts. Drawing on the classical sense-plan-act paradigm in robotics~\cite{fikes1971strips,kortenkamp2016robotic} and its recent extension to language-grounded embodied agents~\cite{bai2026egoactor,shou2026halo}, we decompose SDCL execution into four capability dimensions spanning low-level perception to high-level inference, each targeting distinct judgments required in real chemical experiments.

\paragraph{Actionability.}
Actionability evaluates whether an MLLM can translate laboratory observations into physically actionable affordances and trajectories, which requires identifying operable regions, feasible interactions, and action-relevant motion paths under laboratory constraints.

\paragraph{Knowledge.}
Reliable action presupposes accurate recognition of laboratory entities and awareness of procedural norms. This dimension evaluates whether a model can identify apparatus and reagents, understand their current states (e.g., liquid phase, precipitate formation, color changes), and judge whether an observed setup conforms to standard experimental practice.

\paragraph{Experimentation.}
An embodied brain must also understand experimental operations as procedural processes~\cite{yagi2025finebio}. Experimentation evaluates whether a model can recognize ongoing operations, understand apparatus connections, infer procedural intent, and track step-level progress.

\paragraph{Reasoning.}
Chemical experiments require models to integrate visual evidence, experimental context, and domain knowledge to infer state changes, causal relationships, and experimental outcomes~\cite{xu2025expvid}. This dimension evaluates whether a model can reason over temporal observations of varying lengths and interpret experimental results from accumulated evidence, which is essential for closing the decision loop in autonomous SDCL workflows.

\subsection{Benchmark Construction}
\label{sec:construction}

\paragraph{Expert-Curated Data Collection.}
All visual data in BEAKER is collected and filtered entirely by domain experts. For images, we survey publicly available chemical laboratory datasets across academic repositories and community platforms, identifying 12 representative sources covering diverse real-world laboratory scenes (see Appendix~\ref{app:data_sources}). Chemistry experts screen 11,128 candidate images and retain 883 that reflect realistic execution contexts: 275 \emph{simple} images featuring a single apparatus or reagent against a clean background, and 608 \emph{complex} images involving cluttered environments, multi-step setups, or active experimenter operations. Two senior embodied experts then verify that all retained images are suitable for embodied task design. Many images adopt first-person or experiment-execution viewpoints, resembling the visual inputs of automated devices in real SDCL execution.

\input{figures/pipeline}

For videos, 334 short clips (4\,s each) are extracted from the egocentric IBM Lab Actions dataset~\cite{gabrieli2025activity}; 200 long experimental recordings (average duration $\sim$9\,min) with accompanying expert-authored transcripts are curated from the Journal of Visualized Experiments (JoVE) Chemistry collection~\cite{jove_chemistry}; and mid-length clips are derived from these long videos via a semantic segmentation pipeline (FFmpeg~\cite{ffmpeg2024} for audio extraction, Whisper~\cite{radford2023whisper} for timestamped transcription, and boundary detection between individual experimental steps), producing 1,753 candidate segments, of which experts retain 200 semantically complete clips (average duration $\sim$50\,s). Whisper only provides speech timestamps for proposing candidate segments; it is not involved in sample selection, question authoring, or annotation, all of which are performed by human experts. In total, the curated data pool comprises 883 images and 734 video clips.

Ten chemistry experts then author detailed free-form captions for every item, describing apparatus, reagent states, ongoing operations, and safety-relevant details, yielding 1,617 expert-authored captions. For mid-length and long videos, the JoVE narration transcripts are not directly applicable to constructing BEAKER's QA data, so experts combine each clip with its transcript to write captions tailored to this~benchmark.

\paragraph{Expert Annotation and QA Design.}\label{sec:annotation}
Recent studies show that benchmarks generated by Large Language Models (LLMs) introduce systematic self-bias~\cite{yuan2025silencer} and that LLM evaluators exhibit self-preference~\cite{panickssery2024selfpreference}. To reduce reliance on model-generated supervision, all benchmark questions and reference answers are authored and reviewed by human domain experts, without AI-assisted question or answer generation. Our annotation team comprises 10 chemistry experts, who ensure scientific correctness and domain accuracy, and 12 embodied experts, who verify operational relevance and feasibility from a robotic execution perspective. Two senior embodied experts coordinate overall quality control; the remaining 10 are each paired with one chemistry expert to form 10 interdisciplinary annotation groups. To ensure balanced task coverage while minimizing redundant annotations, each group receives the full curated data pool along with a \emph{task-type allocation matrix} (see Appendix~\ref{app:allocation_matrix}) specifying its question quota per category. Groups independently select suitable data items and design 150 QA samples each, yielding an initial pool of \textbf{1,500} candidate questions across 11 task types. Because all groups draw from the same pool, visually rich items may be selected by multiple groups; the allocation matrix steers these groups toward \emph{different} task types for the same item, increasing per-item competency diversity rather than redundancy.

To calibrate quality and difficulty, we apply a cross-group blind-review protocol inspired by expert-validation methodologies~\cite{rein2023gpqa}. Each group's 150 questions are distributed to two non-originating groups, whose members independently attempt every question \emph{without} access to the ground-truth answer and evaluate it on four criteria: (i)~scientific correctness, (ii)~embodied operational relevance, (iii)~difficulty calibration, and (iv)~answer unambiguity. Reviewers also flag \emph{semantic duplicates}, i.e., questions from different groups that target the same data item and competency, retaining only the highest-rated variant. Flagged questions enter a consensus adjudication round moderated by the two senior embodied experts, in which the originating and reviewing groups jointly decide whether to revise or discard each item. A question is discarded if it is answerable through surface-level visual cues alone (\emph{e.g.}, ``\emph{What is the color of the rubber-bulb pipette the lab worker is holding?}''), reduces to a trivial binary judgment without diagnostic value, demands excessively specialized knowledge beyond graduate-level chemistry, or lacks grounding in embodied execution contexts. Through this process, 500 questions (33.3\%) are removed, and the remaining \textbf{1,000} constitute the final BEAKER benchmark, drawn from 410 images (152 simple, 258 complex) and 195 short, 125 mid-length, and 80~long~videos.

Fig.~\ref{fig:pipeline} illustrates this pipeline with task examples, and Fig.~\ref{fig:data_statistics} summarizes the resulting statistics. The image tasks (600 samples) include two spatial tasks, Affordance Grounding (AFF) and Trajectory Planning (TRAJ), with 150 samples each annotated using the Computer Vision Annotation Tool (CVAT)~\cite{cvat2024}: AFF follows the point-and-box protocol of RoboBrain~2.0~\cite{robobrain2025robobrain2}, with fine-grained boxes on precise regions such as stopcock tips and pipette nozzles, and TRAJ is annotated as ordered keypoint sequences. They also include five Multiple-Choice Question (MCQ) categories with 60 samples each (20 simple- and 40 complex-image questions): Equipment and Reagent Identification (ERI), Experimental Safety and Standardization (ESS), State Recognition (SR), Experimental Apparatus Setup (EAS), and Experimental Operations (EOP). The video tasks (400 samples) comprise Temporal Sequence Reasoning (TSR; 150 MCQs), evaluated separately on short, mid-length, and long videos as S-TSR, M-TSR, and L-TSR; Short-Video Operation Reasoning (SVOR; 125 MCQs); Mid-Video Task-Step Reasoning (MVTS; 75 MCQs); and Long-Video Experimental Procedure Reasoning (LVPR; 50 fill-in-the-blank questions requiring domain-specific phrases). The entire pipeline, from data screening through captioning, question authoring, and cross-group validation, spans about 1,200 expert-hours. Even for tasks that overlap with general embodied benchmarks, all questions are grounded in chemical laboratory execution contexts, so BEAKER evaluates domain-specific understanding rather than general visual competence.

\input{figures/data}

To quantify cross-group consistency, we compare spatial annotations from the originating group and an independent reviewing group. On the retained AFF questions, mean box Intersection over Union (IoU) reaches 86.9\% and bidirectional point-in-region agreement reaches 100.0\%; on the retained TRAJ questions, the mean trajectory agreement score is 85.6/100 (150 questions per task). Appendix~\ref{app:annotation_agreement} details the metrics and compares candidate and retained annotations.

%% file: figures/pipeline.tex
\begin{figure*}[t]
    \centering
    \includegraphics[width=1\textwidth]{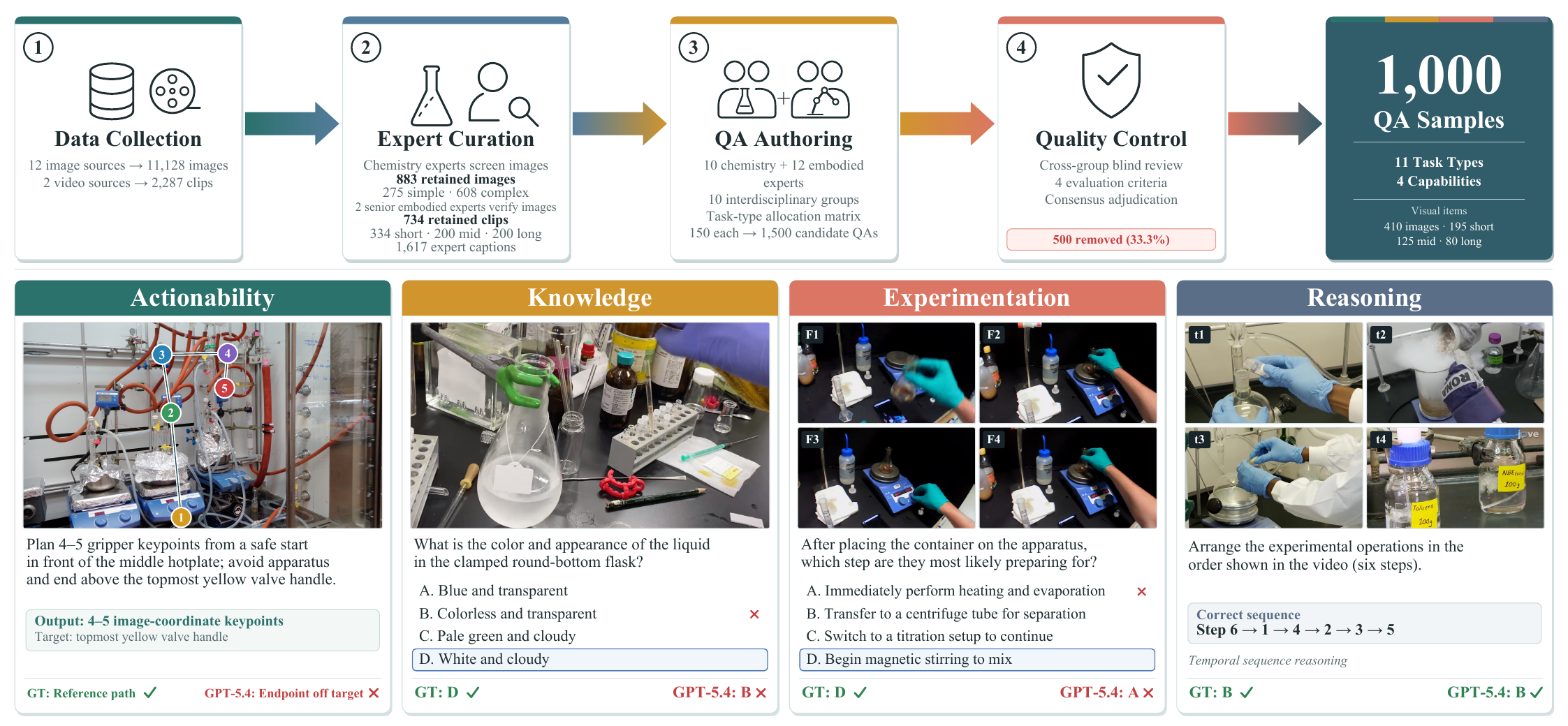}
    \caption{\textbf{Overview of BEAKER construction and task examples.} \textbf{Top:} the four-stage expert-curated pipeline of data collection, expert curation, QA authoring, and quality control. \textbf{Bottom:} examples of the four capability dimensions with the input, question, Ground Truth (GT), and \mbox{GPT-5.4}~\cite{openai2026gpt54} prediction.}
    \label{fig:pipeline}
\end{figure*}

%% file: figures/data.tex
\begin{figure}[t]
    \centering
    \includegraphics[width=1\columnwidth]{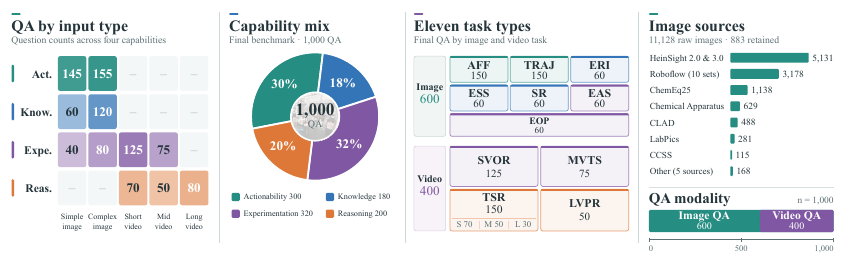}
    \caption{\textbf{Statistics of BEAKER.} Counts by input type, capability, task, data source, and modality.}
    \label{fig:data_statistics}
\end{figure}

%% file: sections/4_experiments.tex
\section{Experiments}
\label{sec:experiments}

\subsection{Benchmarking Procedure}
\label{sec:benchmarking_procedure}

We evaluate 28 mainstream MLLMs on BEAKER, covering general-purpose, science-oriented, chemistry-specialized, and embodied multimodal models, through remote Application Programming Interfaces (APIs) or local deployment on 16 NVIDIA A100 Graphics Processing Units (GPUs). All models are evaluated in a single-turn QA setting with standardized task-specific prompts and output formats (Appendix~\ref{app:prompts}) and scored with task-specific metrics (Appendix~\ref{app:evaluation}). When configurable, the temperature is set to 0.7. All images are input at their original resolution. Short and mid-length videos are represented by 8 and 32 sampled frames, respectively, and long videos by 128 frames arranged into 32 four-frame grids (see Section~\ref{sec:ablation} for a frame sampling ablation). The human baseline averages the scores of three chemistry experts who were not involved in any stage of BEAKER construction, each independently answering all 1{,}000 questions without access to the ground truth.

BEAKER aggregates task scores hierarchically into the four capability dimensions and combines them into the Overall score with expert-defined weights of 0.25, 0.15, 0.35, and 0.25 for Actionability, Knowledge, Experimentation, and Reasoning, respectively. Tasks within each dimension are averaged uniformly, except that Experimentation weights its image-based (EAS, EOP) and video-based (SVOR, MVTS) task groups at a ratio of 2:5. These weights are established through domain-expert consensus to reflect the relative importance of each capability in chemical laboratory execution. All scores are on a 0--100 scale and aggregated before rounding.

\input{table/consolidated_results}

\subsection{Overall Performance}
\label{performance}

\paragraph{Model-Level Insights.}
As shown in Table~\ref{tab:benchmark_consolidated_results}, current MLLMs still exhibit limited embodied understanding in SDCL execution contexts. General-purpose MLLMs achieve the strongest overall performance, while chemistry-specialized and science-oriented MLLMs do not consistently outperform them. Embodied brain models gain on some action-related tasks relative to the similarly sized Qwen3-VL-8B, but these gains remain task-specific. The RynnBrain~\cite{dang2026rynnbrain} variants behave differently: relative to their base model RynnBrain-8B, RynnBrain-CoP-8B markedly improves AFF, whereas RynnBrain-Plan-8B gains on EOP but drops on TRAJ and MVTS, yielding a lower Overall score. Overall rankings are robust to four alternative aggregation schemes (Spearman $\rho \geq 0.980$, Kendall $\tau \geq 0.889$; identical top-5 and top-10 sets), and paired, task-stratified bootstrap 95\% confidence intervals of Overall scores span at most 5.5 points, with 95\% rank ranges of at most seven positions (Appendix~\ref{app:ranking_robustness}); run-to-run variability is analyzed in Appendix~\ref{app:stability}.

\paragraph{Task-Level Analysis.}
Averaged over the 28 models, Actionability is the clearest bottleneck: AFF reaches only 9.7 and TRAJ 34.7, as models must localize small operable parts such as valves, nozzles, handles, and connection interfaces rather than merely recognize objects. Knowledge-oriented image MCQs are also difficult: ERI, ESS, and SR average only 40.0--43.9, against a 30.4\% random-choice baseline for image MCQs (Appendix~\ref{app:mcq_controls}), whereas human experts score at least 93.3 on each. In contrast, models perform better at following ongoing operations in videos (SVOR 65.7, MVTS 56.5), suggesting that procedural context is easier to exploit than precise identification of laboratory apparatus, reagent states, and safety norms. Within TSR, mid-length videos are the hardest (M-TSR 37.1 vs.\ 68.8 for S-TSR and 57.0 for L-TSR; Appendix~\ref{app:tsr_breakdown}), as their step boundaries are less visually salient. Thus, BEAKER locates the main bottlenecks in action grounding and fine-grained domain-specific perception rather than in coarse procedure following; qualitative examples are given in Section~\ref{sec:qualitative}.

\paragraph{Chemistry Experts vs.\ MLLMs.}
Human experts achieve an overall score of 93.3, 34.1 points above the best-performing MLLM, Claude Opus 4.7. The gap widens to 48.0 points for the best embodied brain model, RoboBrain2.5-8B. This difference is critical in chemical laboratory scenarios, which require precise localization of small functional regions and reliable interpretation of procedural operations. Despite recent progress, current MLLMs and embodied models thus remain far from expert-level laboratory understanding.

\paragraph{Specialized vs.\ General Models.}
Table~\ref{tab:specialization_gains_losses} shows selective gains over Qwen3-VL-8B: RoboBrain2.5-8B improves TRAJ and TSR, while Intern-S1-Mini improves LVPR. All three specialized models still score lower overall. Differences in backbones and training recipes prevent attributing these gaps to specialization alone; Section~\ref{sec:failure_attribution} analyzes likely causes.

\input{table/specialization_gains_losses}

\subsection{Capability Profile Analysis}
\label{sec:capability_profile}

\input{table/Performance_different}

Table~\ref{tab:capability_profile} presents the capability profiles of representative MLLMs (Table~\ref{tab:capability_all} in Appendix~\ref{app:capability_all} covers all 28 models) and reveals a clear imbalance. Averaged over the 13 representative models shown, Actionability obtains the lowest score, indicating that MLLMs still struggle to ground laboratory observations into operable regions, feasible interactions, and action-relevant trajectories. In contrast, Experimentation and Reasoning achieve higher averages, suggesting that models are more capable of recognizing experimental operations, following procedural cues, and interpreting temporal evidence than of performing fine-grained action grounding. Model-level patterns also differ: Claude Opus 4.7 achieves the strongest overall performance and the best Actionability score; Gemini 3.1 Pro Preview and Qwen3.5-397B-A17B perform strongly on Experimentation; and Grok-4.20-Reasoning leads on Reasoning. Section~\ref{sec:question_formats} further shows that most models score markedly lower on coordinate-based tasks than on MCQs. BEAKER thus reveals not only whether a model performs well overall, but also which embodied capability remains the bottleneck for SDCL automation.

\subsection{Performance Across Question Formats}
\label{sec:question_formats}

BEAKER employs three distinct answer formats---Coordinate (Coord.) QA (AFF and TRAJ, 300 samples), MCQ (eight task types, 650 samples), and Fill-in-the-Blank (FIB) QA (LVPR, 50 samples)---each assessing different aspects of laboratory understanding and execution. Table~\ref{tab:question_format_analysis} compares five representative models across these task groups and reports pairwise performance ratios to characterize their relative performance. MCQ scores are calculated as the unweighted macro-average across the eight MCQ task types. Since these task groups differ in content, difficulty, and evaluation metrics, the ratios reflect task-group performance differences rather than isolated effects of answer formats.

\input{table_appendix/question_formats}

The most striking finding is the \emph{substantial performance gap in coordinate-based tasks}. All five models exhibit a Coord./MCQ ratio of $\leq$0.7, indicating markedly lower performance on Coord.\ QA than on MCQs. For instance, Qwen3.5-397B-A17B achieves 62.2 on MCQs but only 20.6 on Coord.\ QA (Coord./MCQ\,=\,0.3), while RoboBrain2.5-8B similarly scores 52.2 and 16.6, respectively. This disparity highlights a critical challenge for embodied laboratory intelligence: beyond recognizing experimental scenes and selecting predefined answers, models must generate precise spatial information, such as operable regions and trajectory waypoints, to support physical execution. In real SDCL environments, autonomous systems cannot rely on predefined MCQ options but must produce actionable outputs from visual observations. Therefore, coordinate-based evaluation provides evidence of spatial actionability that conventional MCQ-based assessments alone cannot capture.

Notably, even the two models with the highest Coord.\ QA scores, Claude Opus 4.7 (41.7) and GPT-5.5 (40.0), reach a Coord./MCQ ratio of only 0.7, indicating that the coordinate-based gap persists among the strongest general-purpose models. Moreover, RoboBrain2.5-8B---an embodied brain model designed for spatial reasoning---achieves a Coord./MCQ ratio of only 0.3, indicating that strong performance in general embodied scenarios does not necessarily translate into equally strong fine-grained spatial grounding in chemical laboratory environments. These patterns reveal an important capability imbalance across current MLLMs and reinforce the need for domain-specific evaluation of embodied intelligence in SDCLs.

FIB QA exhibits a different performance pattern, with FIB/MCQ ratios approaching 1.0 for several models, indicating comparable performance between LVPR and the broader MCQ task group. Unlike coordinate-based tasks, FIB QA primarily requires models to generate domain-specific experimental concepts and procedural outcomes from visual and temporal evidence. Although differences in task content and evaluation metrics prevent attributing these results solely to answer formats, the comparison highlights the distinct challenges of spatial action grounding and experimental knowledge reasoning. Together, these findings demonstrate the importance of evaluating complementary capabilities beyond conventional MCQ-based assessments for reliable SDCL automation.

\subsection{Qualitative Analysis}
\label{sec:qualitative}

\begin{figure}[t]
  \centering
  \includegraphics[width=\textwidth]{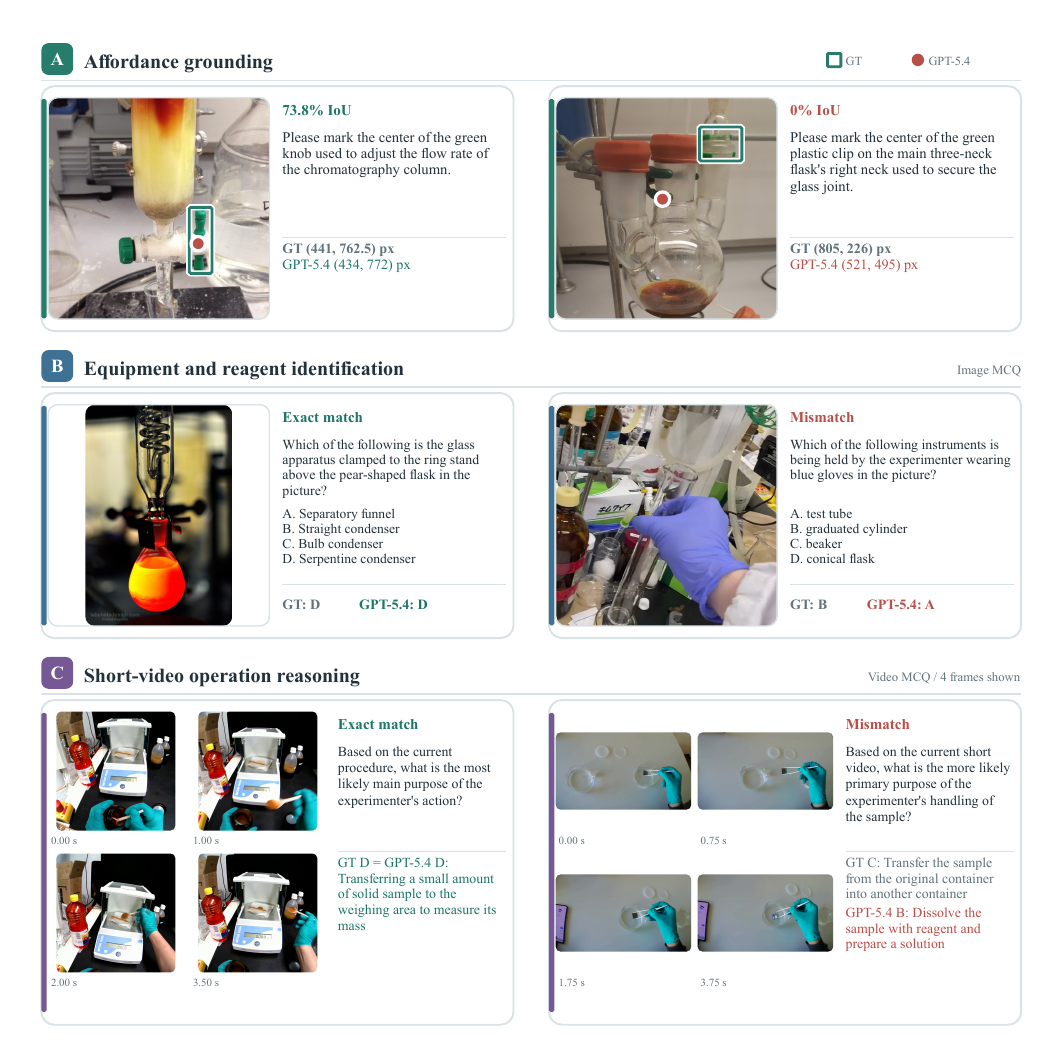}
  \caption{Qualitative examples for three representative task types: \textbf{Affordance Grounding (AFF)} (Actionability, image), \textbf{Equipment and Reagent Identification (ERI)} (Knowledge, image MCQ), and \textbf{Short-Video Operation Reasoning (SVOR)} (Experimentation, video MCQ). Left panels show correct GPT-5.4~\cite{openai2026gpt54} predictions; right panels show incorrect ones. GT: Ground Truth.}
  \label{fig:qa_examples}
\end{figure}

\paragraph{Qualitative Examples.}
Fig.~\ref{fig:qa_examples} presents representative correct and incorrect predictions made by GPT-5.4 on three task types spanning different capability dimensions in BEAKER: AFF for \emph{Actionability}, ERI for \emph{Knowledge}, and SVOR for \emph{Experimentation}. For each task type, the left panel shows a case where GPT-5.4 succeeds and the right panel a case where it fails. The header of each panel reports the outcome, i.e., the IoU score for AFF (73.8\% vs.\ 0\%) and ``Exact match'' or ``Mismatch'' with the ground-truth option for the two MCQ tasks, with successful and failed cases highlighted in green and red, respectively. These examples illustrate the diverse challenges posed by BEAKER across both image- and video-based inputs. For the video examples, 2$\times$2 grids of keyframes sampled from the original clips are shown for compact visualization; these grids are used only for illustration and do not alter the actual evaluation protocol.

\paragraph{Self-Explanation Case Study.}
We conduct a self-explanation case study on LVPR (Fig.~\ref{fig:case_study}), as long videos require models to connect temporally distributed operations, visual evidence, and domain knowledge. For these questions, GPT-5.4 is prompted to state the basis for each answer when responding; these explanations are not scored. In the correct case, GPT-5.4 links the observed operations to the required answers: tissue pre-digestion, sonication, and dilution indicate Inductively Coupled Plasma Mass Spectrometry (ICP-MS) analysis; platinum standards and the concentration--signal plot support calibration-curve construction; and repeated quality-control samples indicate drift monitoring. In the incorrect case, the model captures the coarse biological effect but overgeneralizes the evidence, replacing the required component-level answers with broader mechanistic descriptions such as antioxidant defense. Thus, GPT-5.4 can extract procedural evidence from long videos yet may fail to map observations to the required fine-grained chemical concepts, revealing a gap between procedural understanding and precise experimental knowledge grounding.

\input{figures/Self-explanation}

\subsection{Ablation Studies}
\label{sec:ablation}

\paragraph{Input Modality Ablation for Image Understanding.}
Fig.~\ref{fig:image_modality_ablation} compares image-plus-text and text-only inputs on 300 image MCQs. Removing images reduces the observed average accuracy for all eight models on the evaluated questions, demonstrating the contribution of visual information under the tested settings. General-purpose and embodied models benefit substantially, whereas the science-oriented Intern-S1 and the chemistry-specialized ChemVLM-8B benefit little, suggesting limited use of visual evidence. Random-choice and majority-option baselines, along with answer-position distributions, are provided in Appendix~\ref{app:mcq_controls}.

\input{figures/visual_dependence_dumbbell}

\paragraph{Input Modality Ablation for Video Reasoning.}
Fig.~\ref{fig:modality_ablation} compares three input settings on mid- and long-video tasks: video-plus-question, caption-plus-question, and caption-plus-video-plus-question. Adding captions substantially improves performance over video inputs alone, especially on M-TSR, L-TSR, and LVPR, as expert-written captions provide procedural details that help models answer the questions. The largest gain appears on M-TSR, where captions roughly double the scores of all three models, suggesting that the difficulty of mid-length videos lies mainly in extracting step boundaries from frames rather than in ordering described steps. The gap between video-only and caption-based settings suggests that current MLLMs still struggle to extract fine-grained operations, temporal dependencies, and chemistry-specific procedural cues directly from multi-frame laboratory videos.

\input{figures/modality_bar}

\paragraph{Frame Sampling Ablation.}
We investigate the sensitivity of video-based BEAKER tasks to frame sampling density by evaluating Qwen3.5-397B-A17B under three settings: fewer frames (4/16/64), standard frames (8/32/128, used in all main experiments), and more frames (16/48/192) for short/mid/long videos, respectively. In all settings, each video is divided into $K$ equal-length segments, where $K$ is the number of sampled frames, and the frame at the midpoint of each segment is extracted. For long videos, every four consecutive frames are arranged in temporal order into a $2\times2$ image grid before being fed to the model (e.g., 128 frames yield 32 grids).

\input{table_appendix/Different_frame}

As shown in Table~\ref{tab:frame_sampling_ablation}, S-TSR improves from 82.9 to 88.6 as the number of frames increases from 4 to 16, whereas SVOR exhibits a non-monotonic trend (80.8, 79.2, and 80.8). For mid-length videos, M-TSR scores 54.0, 48.0, and 52.0 under fewer, standard, and more frames, respectively, while MVTS achieves its highest score of 65.3 under the standard setting. For long videos, L-TSR improves monotonically with more frames (66.7, 70.0, and 73.3), whereas LVPR peaks at the standard setting (64.7) before declining to 55.3 with more frames. Overall, these results indicate that increasing frame sampling density does not consistently improve performance across video reasoning tasks. Redundant visual information may contribute to this pattern, although the experiment does not isolate its cause. These findings support the standard setting (8/32/128) as a practical configuration and suggest that, although denser sampling helps temporal ordering in long videos, increasing frame count alone does not reliably improve long-horizon procedure reasoning.

%% file: table/consolidated_results.tex
\definecolor{rank1}{HTML}{0000CD} % 1st: Medium Blue
\definecolor{rank2}{HTML}{B22222} % 2nd: Firebrick Red
\definecolor{rank3}{HTML}{006400} % 3rd: Dark Green

\begin{table*}[t]
\centering
\caption{\textbf{Consolidated results on BEAKER.} Task scores are grouped by the four capability dimensions, and Overall aggregates them as described in Section~\ref{sec:benchmarking_procedure}. Model rankings within each column are color-coded: \textcolor{rank1}{\textbf{1st}}, \textcolor{rank2}{\textbf{2nd}}, and \textcolor{rank3}{\textbf{3rd}}; tied scores share the same rank.}
\label{tab:benchmark_consolidated_results}
\vspace{2mm}
\resizebox{\textwidth}{!}{
\begin{tabular}{l cc ccc cccc cc c}
\toprule
\multirow{2}{*}{Model} 
& \multicolumn{2}{c}{\textbf{Actionability}} 
& \multicolumn{3}{c}{\textbf{Knowledge}} 
& \multicolumn{4}{c}{\textbf{Experimentation}} 
& \multicolumn{2}{c}{\textbf{Reasoning}} 
& \multirow{2}{*}{\textbf{Overall}} \\
\cmidrule(lr){2-3} 
\cmidrule(lr){4-6} 
\cmidrule(lr){7-10} 
\cmidrule(lr){11-12}
& AFF$\uparrow$ 
& TRAJ$\uparrow$ 
& ERI$\uparrow$ 
& ESS$\uparrow$ 
& SR$\uparrow$ 
& EAS$\uparrow$ 
& EOP$\uparrow$ 
& SVOR$\uparrow$ 
& MVTS$\uparrow$ 
& TSR$\uparrow$ 
& LVPR$\uparrow$ 
& \\
\midrule

Human 
& 88.0 
& 91.0 
& 96.7 
& 95.0 
& 93.3 
& 93.3 
& 96.7 
& 94.4 
& 92.0 
& 95.0 
& 96.0 
& 93.3 \\

\midrule

\multicolumn{13}{c}{\textbf{Chemistry-specialized Multimodal Models}} \\
\midrule
\rowcolor{gray!10} Open-source Models & & & & & & & & & & & & \\
\midrule

ChemVLM-8B~\cite{li2025chemvlm}
& 2.1
& 21.0
& 33.3
& 40.0
& 21.7
& 36.7
& 33.3
& 36.0
& 40.0
& 37.9
& 24.0
& 28.4 \\

ChemVLM-26B~\cite{li2025chemvlm}
& 3.1
& 27.0
& 23.3
& 25.0
& 31.7
& 36.7
& 33.3
& 37.6
& 42.7
& 42.4
& 28.0
& 30.1 \\

\midrule

\multicolumn{13}{c}{\textbf{Science-oriented Multimodal Models}} \\
\midrule
\rowcolor{gray!10} Open-source Models & & & & & & & & & & & & \\
\midrule

Intern-S1~\cite{bai2025interns1}
& 3.4
& 24.6
& 40.0
& 25.0
& 48.3
& 33.3
& 48.3
& 61.6
& 60.0
& 41.4
& 54.8
& 40.5 \\

Intern-S1-Mini~\cite{bai2025interns1}
& 1.7
& 22.2
& 43.3
& 38.3
& 43.3
& 30.0
& 41.7
& 46.4
& 56.0
& 24.0
& 52.7
& 35.2 \\

Intern-S1-Pro~\cite{zou2026interns1pro}
& 2.9
& 18.0
& 45.0
& 38.3
& 51.7
& 30.0
& \textcolor{rank1}{\textbf{71.7}}
& 72.0
& \textcolor{rank2}{\textbf{65.3}}
& 37.0
& 62.2
& 44.0 \\

\midrule

\multicolumn{13}{c}{\textbf{Embodied Brain Models}} \\
\midrule
\rowcolor{gray!10} Open-source Models & & & & & & & & & & & & \\
\midrule

ACE-Brain-0-8B~\cite{gong2026ace}
& 2.4
& 23.0
& 33.3
& \textcolor{rank1}{\textbf{76.7}}
& 31.7
& \textcolor{rank2}{\textbf{63.3}}
& 55.0
& 68.8
& \textcolor{rank2}{\textbf{65.3}}
& 54.3
& 40.0
& 44.7 \\

Pelican-VL-1.0-7B~\cite{zhang2025pelicanvl}
& 10.9
& 24.2
& 26.7
& 41.7
& 30.0
& \textcolor{rank3}{\textbf{55.0}}
& \textcolor{rank3}{\textbf{63.3}}
& 50.4
& 48.0
& 28.5
& 32.8
& 35.2 \\

RoboBrain2.5-8B~\cite{robobrain2026robobrain25}
& 2.3
& 30.9
& 41.7
& 35.0
& 48.3
& 43.3
& 58.3
& 74.4
& 61.3
& 55.3
& 47.5
& 45.3 \\

RynnBrain-8B~\cite{dang2026rynnbrain}
& 3.2
& 40.7
& 28.3
& 45.0
& 28.3
& 45.0
& 43.3
& 52.8
& 46.7
& 47.3
& 47.3
& 39.2 \\

RynnBrain-CoP-8B~\cite{dang2026rynnbrain}
& \textcolor{rank3}{\textbf{28.6}}
& 38.6
& 25.0
& 41.7
& 30.0
& 28.3
& 41.7
& 56.8
& 50.7
& 53.8
& 47.7
& 42.9 \\

RynnBrain-Plan-8B~\cite{dang2026rynnbrain}
& 7.0
& 28.0
& 31.7
& 48.3
& 26.7
& 40.0
& 60.0
& 52.0
& 37.3
& 47.2
& 45.3
& 37.4 \\

MiMo-Embodied-7B~\cite{mimo2025embodied}
& 9.8
& 28.9
& 23.3
& 38.3
& 43.3
& 33.3
& 51.7
& 57.6
& 48.0
& 45.9
& 48.0
& 39.3 \\

\midrule

\multicolumn{13}{c}{\textbf{General Multimodal Large Language Models}} \\
\midrule
\rowcolor{gray!10} Closed-source Models & & & & & & & & & & & & \\

GPT-5.5~\cite{openai2026gpt55}
& \textcolor{rank2}{\textbf{32.7}}
& \textcolor{rank2}{\textbf{47.2}}
& \textcolor{rank2}{\textbf{55.0}}
& 28.3
& \textcolor{rank1}{\textbf{61.7}}
& \textcolor{rank2}{\textbf{63.3}}
& \textcolor{rank1}{\textbf{71.7}}
& 72.0
& 49.3
& 65.0
& \textcolor{rank3}{\textbf{64.7}}
& 55.4 \\

GPT-5.4~\cite{openai2026gpt54}
& 26.8
& 46.1
& 48.3
& \textcolor{rank3}{\textbf{58.3}}
& 48.3
& 53.3
& \textcolor{rank1}{\textbf{71.7}}
& \textcolor{rank2}{\textbf{80.8}}
& 57.3
& \textcolor{rank3}{\textbf{70.3}}
& 64.2
& \textcolor{rank2}{\textbf{57.2}} \\

Claude Opus 4.7~\cite{anthropic2026opus47}
& \textcolor{rank1}{\textbf{34.5}}
& \textcolor{rank1}{\textbf{48.9}}
& 45.0
& 55.0
& \textcolor{rank1}{\textbf{61.7}}
& \textcolor{rank1}{\textbf{65.0}}
& \textcolor{rank3}{\textbf{63.3}}
& \textcolor{rank2}{\textbf{80.8}}
& \textcolor{rank3}{\textbf{64.0}}
& 67.2
& 62.2
& \textcolor{rank1}{\textbf{59.2}} \\

Claude Sonnet 4.6~\cite{anthropic2026sonnet46}
& 18.3
& \textcolor{rank3}{\textbf{46.3}}
& 41.7
& 38.3
& 45.0
& 45.0
& 56.7
& 76.8
& \textcolor{rank3}{\textbf{64.0}}
& 65.8
& 54.8
& 52.1 \\

Gemini 2.5 Pro~\cite{google2026gemini25pro}
& 7.4
& 44.4
& 43.3
& 23.3
& 53.3
& 20.0
& 45.0
& 47.2
& 40.0
& 46.3
& \textcolor{rank2}{\textbf{67.0}}
& 40.8 \\

Gemini 3.1 Pro Preview~\cite{google2026gemini31pro}
& 4.7
& 29.6
& \textcolor{rank1}{\textbf{66.7}}
& 33.3
& \textcolor{rank2}{\textbf{60.0}}
& \textcolor{rank1}{\textbf{65.0}}
& \textcolor{rank2}{\textbf{70.0}}
& \textcolor{rank1}{\textbf{82.4}}
& \textcolor{rank3}{\textbf{64.0}}
& \textcolor{rank1}{\textbf{85.9}}
& 59.2
& \textcolor{rank3}{\textbf{55.5}} \\

Grok-4.20-Non-Reasoning~\cite{xai2026grok420}
& 8.7
& 39.0
& 31.7
& 38.3
& 31.7
& 25.0
& 56.7
& 65.6
& 53.3
& 63.6
& 54.0
& 44.7 \\

Grok-4.20-Reasoning~\cite{xai2026grok420}
& 9.7
& 43.3
& 38.3
& 33.3
& \textcolor{rank1}{\textbf{61.7}}
& 31.7
& 56.7
& 71.2
& 60.0
& \textcolor{rank2}{\textbf{83.1}}
& \textcolor{rank1}{\textbf{67.7}}
& 52.9 \\

MiMo-V2-Omni~\cite{mimov2omni_2026}
& 9.2
& 42.6
& 38.3
& 41.7
& 40.0
& 36.7
& 58.3
& 76.0
& \textcolor{rank1}{\textbf{66.7}}
& 51.9
& 58.2
& 48.8 \\

Qwen3.6-Plus~\cite{qwen36_2026}
& 4.9
& 41.1
& 41.7
& 36.7
& 41.7
& 43.3
& \textcolor{rank1}{\textbf{71.7}}
& 76.8
& 62.7
& 68.4
& 58.3
& 50.8 \\

\midrule
\rowcolor{gray!10} Open-source Models & & & & & & & & & & & & \\

Kimi-K2.5~\cite{kimi_k25}
& 6.7
& 42.5
& 43.3
& \textcolor{rank2}{\textbf{63.3}}
& 43.3
& 46.7
& \textcolor{rank1}{\textbf{71.7}}
& \textcolor{rank3}{\textbf{79.2}}
& 62.7
& 60.3
& 59.3
& 52.2 \\

Kimi-K2.6~\cite{kimik26_2026}
& 5.9
& 41.8
& 46.7
& 41.7
& 45.0
& 46.7
& \textcolor{rank1}{\textbf{71.7}}
& \textcolor{rank3}{\textbf{79.2}}
& \textcolor{rank3}{\textbf{64.0}}
& 64.4
& 60.3
& 52.0 \\

MiMo-V2.5~\cite{mimov25_2026}
& 11.3
& 40.1
& 45.0
& 41.7
& 45.0
& 36.7
& 61.7
& 72.0
& 61.3
& 51.0
& 57.8
& 48.2 \\

Qwen3-VL-8B~\cite{bai2025qwen3}
& 4.1
& 26.8
& 45.0
& 41.7
& \textcolor{rank3}{\textbf{56.7}}
& 45.0
& 60.0
& 75.2
& \textcolor{rank1}{\textbf{66.7}}
& 54.2
& 50.2
& 47.1 \\

InternVL3.5-241B-A28B~\cite{wang2025internvl35}
& 5.2
& 28.1
& 45.0
& 28.3
& 53.3
& 38.3
& 43.3
& 59.2
& 58.7
& 40.5
& 59.5
& 41.8 \\

Qwen3.5-397B-A17B~\cite{qwen35_397b}
& 3.8
& 37.3
& \textcolor{rank3}{\textbf{50.0}}
& 53.3
& 45.0
& \textcolor{rank1}{\textbf{65.0}}
& \textcolor{rank1}{\textbf{71.7}}
& \textcolor{rank3}{\textbf{79.2}}
& \textcolor{rank2}{\textbf{65.3}}
& 67.9
& \textcolor{rank3}{\textbf{64.7}}
& 54.0 \\

\bottomrule
\end{tabular}
}

\vspace{0.5mm}
\begin{minipage}{\textwidth}
\tiny
\setlength{\baselineskip}{6pt}
\noindent\textbf{Note:} AFF: Affordance Grounding; TRAJ: Trajectory Planning; ERI: Equipment and Reagent Identification; ESS: Experimental Safety and Standardization; SR: State Recognition; EAS: Experimental Apparatus Setup; EOP: Experimental Operations; SVOR: Short-Video Operation Reasoning; MVTS: Mid-Video Task-Step Reasoning; TSR: Temporal Sequence Reasoning; LVPR: Long-Video Experimental Procedure Reasoning.
\end{minipage}
\end{table*}    

%% file: table/specialization_gains_losses.tex
\begin{table}[htbp]
\centering
\caption{\textbf{Specialization gains and losses relative to Qwen3-VL-8B}~\cite{bai2025qwen3}\textbf{.} Task cells report score differences with paired wins/losses below; ties are omitted. Avg.\ reports the difference in unweighted capability means (and thus differs from the 2:5-weighted Experimentation score in Table~\ref{tab:capability_all}). Positive and negative differences are shown in green and red, respectively; shaded bold cells summarize capability and Overall differences.}
\label{tab:specialization_gains_losses}
\vspace{1mm}

{\scriptsize
\renewcommand{\arraystretch}{1.10}
\setlength{\tabcolsep}{3pt}

% Vertically center two-line task cells relative to single-line cells.
\renewcommand{\shortstack}[1]{%
  \begin{tabular}[c]{@{}c@{}}#1\end{tabular}%
}

\resizebox{\linewidth}{!}{%
\begin{tabular}{l|rrr|rrrr|rrrrr|rrr|r}
\toprule
\multirow{2}{*}{\textbf{Model}}
& \multicolumn{3}{c|}{\textbf{Actionability}}
& \multicolumn{4}{c|}{\textbf{Knowledge}}
& \multicolumn{5}{c|}{\textbf{Experimentation}}
& \multicolumn{3}{c|}{\textbf{Reasoning}}
& \multirow{2}{*}{\textbf{Overall}} \\
\cmidrule(lr){2-4}
\cmidrule(lr){5-8}
\cmidrule(lr){9-13}
\cmidrule(lr){14-16}
& \textbf{AFF} & \textbf{TRAJ} & \textbf{Avg.}
& \textbf{ERI} & \textbf{ESS} & \textbf{SR} & \textbf{Avg.}
& \textbf{EAS} & \textbf{EOP} & \textbf{SVOR} & \textbf{MVTS} & \textbf{Avg.}
& \textbf{TSR} & \textbf{LVPR} & \textbf{Avg.}
& \\
\midrule

ChemVLM-8B~\cite{li2025chemvlm}
& \shortstack{\textcolor{rank2}{-2.00}\\(13/16)}
& \shortstack{\textcolor{rank2}{-5.78}\\(42/108)}
& \cellcolor{gray!10}\textcolor{rank2}{\textbf{-3.89}}
& \shortstack{\textcolor{rank2}{-11.67}\\(10/17)}
& \shortstack{\textcolor{rank2}{-1.67}\\(12/13)}
& \shortstack{\textcolor{rank2}{-35.00}\\(3/24)}
& \cellcolor{gray!10}\textcolor{rank2}{\textbf{-16.11}}
& \shortstack{\textcolor{rank2}{-8.33}\\(11/16)}
& \shortstack{\textcolor{rank2}{-26.67}\\(6/22)}
& \shortstack{\textcolor{rank2}{-39.20}\\(8/57)}
& \shortstack{\textcolor{rank2}{-26.67}\\(9/29)}
& \cellcolor{gray!10}\textcolor{rank2}{\textbf{-25.22}}
& \shortstack{\textcolor{rank2}{-16.25}\\(16/44)}
& \shortstack{\textcolor{rank2}{-26.17}\\(5/30)}
& \cellcolor{gray!10}\textcolor{rank2}{\textbf{-21.21}}
& \cellcolor{gray!10}\textcolor{rank2}{\textbf{-18.67}} \\
\addlinespace[3pt]

Intern-S1-Mini~\cite{bai2025interns1}
& \shortstack{\textcolor{rank2}{-2.33}\\(7/17)}
& \shortstack{\textcolor{rank2}{-4.63}\\(50/100)}
& \cellcolor{gray!10}\textcolor{rank2}{\textbf{-3.48}}
& \shortstack{\textcolor{rank2}{-1.67}\\(13/14)}
& \shortstack{\textcolor{rank2}{-3.33}\\(9/11)}
& \shortstack{\textcolor{rank2}{-13.33}\\(5/13)}
& \cellcolor{gray!10}\textcolor{rank2}{\textbf{-6.11}}
& \shortstack{\textcolor{rank2}{-15.00}\\(10/19)}
& \shortstack{\textcolor{rank2}{-18.33}\\(11/22)}
& \shortstack{\textcolor{rank2}{-28.80}\\(7/43)}
& \shortstack{\textcolor{rank2}{-10.67}\\(5/13)}
& \cellcolor{gray!10}\textcolor{rank2}{\textbf{-18.20}}
& \shortstack{\textcolor{rank2}{-30.16}\\(15/57)}
& \shortstack{\textcolor{rank3}{+2.50}\\(16/14)}
& \cellcolor{gray!10}\textcolor{rank2}{\textbf{-13.83}}
& \cellcolor{gray!10}\textcolor{rank2}{\textbf{-11.84}} \\
\addlinespace[3pt]

RoboBrain2.5-8B~\cite{robobrain2026robobrain25}
& \shortstack{\textcolor{rank2}{-1.81}\\(8/11)}
& \shortstack{\textcolor{rank3}{+4.08}\\(80/70)}
& \cellcolor{gray!10}\textcolor{rank3}{\textbf{+1.14}}
& \shortstack{\textcolor{rank2}{-3.33}\\(4/6)}
& \shortstack{\textcolor{rank2}{-6.67}\\(5/9)}
& \shortstack{\textcolor{rank2}{-8.33}\\(3/8)}
& \cellcolor{gray!10}\textcolor{rank2}{\textbf{-6.11}}
& \shortstack{\textcolor{rank2}{-1.67}\\(8/9)}
& \shortstack{\textcolor{rank2}{-1.67}\\(6/7)}
& \shortstack{\textcolor{rank2}{-0.80}\\(6/7)}
& \shortstack{\textcolor{rank2}{-5.33}\\(5/9)}
& \cellcolor{gray!10}\textcolor{rank2}{\textbf{-2.37}}
& \shortstack{\textcolor{rank3}{+1.11}\\(14/13)}
& \shortstack{\textcolor{rank2}{-2.67}\\(5/7)}
& \cellcolor{gray!10}\textcolor{rank2}{\textbf{-0.78}}
& \cellcolor{gray!10}\textcolor{rank2}{\textbf{-1.76}} \\

\bottomrule
\end{tabular}%
}
}

\end{table}

%% file: table/Performance_different.tex
\definecolor{BEAKERPurple}{HTML}{8B7BBE}
\definecolor{BEAKERLightGray}{HTML}{F4F1F8}

\newcommand{\bpcell}[2]{\cellcolor{BEAKERPurple!#2}{#1}}
\newcommand{\bpcellbest}[2]{\cellcolor{BEAKERPurple!#2}{\textbf{#1}}}
\newcommand{\rotmodel}[1]{\rotatebox[origin=c]{75}{#1}}
\newcommand{\bpgroup}[3]{\multicolumn{#1}{c}{\textcolor{#2!85!black}{\textbf{#3}}}}

\begin{table*}[t!]
\centering
\caption{\textbf{Capability profile of representative MLLMs on BEAKER.} Avg.\ is averaged over the 13 models shown (all 28 in Table~\ref{tab:capability_all}). Shading scales with the score; the best result in each row is in bold.}
\label{tab:capability_profile}
\vspace{1mm}
\footnotesize
\renewcommand{\arraystretch}{1.12}
\setlength{\tabcolsep}{3.5pt}
\resizebox{\textwidth}{!}{%
\begin{tabular}{@{}l ccccccccccccc@{\hspace{9pt}}c@{}}
\toprule
& \bpgroup{1}{bkCatChem}{Chem.}
& \bpgroup{1}{bkCatSci}{Sci.}
& \bpgroup{3}{bkCatEmb}{Embodied Brain}
& \bpgroup{4}{bkCatGen}{General (Closed-Source)}
& \bpgroup{4}{bkCatGen}{General (Open-Source)}
& \\
\arrayrulecolor{bkRule}\cmidrule(lr){2-2}\cmidrule(lr){3-3}\cmidrule(lr){4-6}\cmidrule(lr){7-10}\cmidrule(lr){11-14}\arrayrulecolor{black}
\raisebox{-1.6em}{\textbf{Capability}}
& \rotmodel{ChemVLM-8B~\cite{li2025chemvlm}}
& \rotmodel{Intern-S1-Pro~\cite{zou2026interns1pro}}
& \rotmodel{ACE-Brain-0-8B~\cite{gong2026ace}}
& \rotmodel{RoboBrain2.5-8B~\cite{robobrain2026robobrain25}}
& \rotmodel{RynnBrain-8B~\cite{dang2026rynnbrain}}
& \rotmodel{GPT-5.5~\cite{openai2026gpt55}}
& \rotmodel{Claude Opus 4.7~\cite{anthropic2026opus47}}
& \rotmodel{Gemini 3.1 Pro Preview~\cite{google2026gemini31pro}}
& \rotmodel{Grok-4.20-Reasoning~\cite{xai2026grok420}}
& \rotmodel{Kimi-K2.6~\cite{kimik26_2026}}
& \rotmodel{MiMo-V2.5~\cite{mimov25_2026}}
& \rotmodel{InternVL3.5-241B-A28B~\cite{wang2025internvl35}}
& \rotmodel{Qwen3.5-397B-A17B~\cite{qwen35_397b}}
& \rotmodel{\textbf{Avg.}} \\
\midrule
\bkdot{bkAct}\enspace Actionability
& \bpcell{11.6}{7}
& \bpcell{10.5}{6}
& \bpcell{12.7}{8}
& \bpcell{16.6}{10}
& \bpcell{21.9}{13}
& \bpcell{40.0}{24}
& \bpcellbest{41.7}{25}
& \bpcell{17.2}{10}
& \bpcell{26.5}{16}
& \bpcell{23.8}{14}
& \bpcell{25.7}{15}
& \bpcell{16.6}{10}
& \bpcell{20.6}{12}
& \bpcell{21.9}{13} \\

\bkdot{bkKnow}\enspace Knowledge
& \bpcell{31.7}{19}
& \bpcell{45.0}{27}
& \bpcell{47.2}{28}
& \bpcell{41.7}{25}
& \bpcell{33.9}{20}
& \bpcell{48.3}{29}
& \bpcellbest{53.9}{32}
& \bpcell{53.3}{32}
& \bpcell{44.4}{27}
& \bpcell{44.4}{27}
& \bpcell{43.9}{26}
& \bpcell{42.2}{25}
& \bpcell{49.4}{30}
& \bpcell{44.6}{27} \\

\bkdot{bkExp}\enspace Experimentation
& \bpcell{37.1}{22}
& \bpcell{63.6}{38}
& \bpcell{64.8}{39}
& \bpcell{63.0}{38}
& \bpcell{48.1}{29}
& \bpcell{62.6}{38}
& \bpcell{70.0}{42}
& \bpcellbest{71.6}{43}
& \bpcell{59.5}{36}
& \bpcell{68.0}{41}
& \bpcell{61.7}{37}
& \bpcell{53.8}{32}
& \bpcell{71.1}{43}
& \bpcell{61.2}{37} \\

\bkdot{bkReas}\enspace Reasoning
& \bpcell{31.0}{19}
& \bpcell{49.6}{30}
& \bpcell{47.2}{28}
& \bpcell{51.4}{31}
& \bpcell{47.3}{28}
& \bpcell{64.8}{39}
& \bpcell{64.7}{39}
& \bpcell{72.6}{44}
& \bpcellbest{75.4}{45}
& \bpcell{62.4}{37}
& \bpcell{54.4}{33}
& \bpcell{50.0}{30}
& \bpcell{66.3}{40}
& \bpcell{56.7}{34} \\

\midrule
\rowcolor{BEAKERLightGray}
\textbf{Overall}
& \bpcell{28.4}{17}
& \bpcell{44.0}{26}
& \bpcell{44.7}{27}
& \bpcell{45.3}{27}
& \bpcell{39.2}{24}
& \bpcell{55.4}{33}
& \bpcellbest{59.2}{36}
& \bpcell{55.5}{33}
& \bpcell{52.9}{32}
& \bpcell{52.0}{31}
& \bpcell{48.2}{29}
& \bpcell{41.8}{25}
& \bpcell{54.0}{32}
& \bpcell{47.7}{29} \\

\bottomrule
\end{tabular}%
}
\end{table*}

%% file: table_appendix/question_formats.tex
\begin{table}[t]
\centering
\caption{\textbf{Performance across question-format task groups.}
We compare representative models across three BEAKER QA task groups: Coordinate (Coord.) QA, MCQ, and Fill-in-the-Blank (FIB) QA. MCQ scores are computed as the unweighted macro-average across the eight MCQ task types. The last three columns report pairwise performance ratios between task groups, reflecting relative performance rather than format-specific robustness. Score cells are shaded by value, and ratio cells by their distance from 1.0.}
\label{tab:question_format_analysis}
\small
\setlength{\tabcolsep}{4pt}
\bksetup{1.3}
\begin{tabularx}{\linewidth}{@{\hspace{5pt}}X ccc ccc@{\hspace{5pt}}}
\bktoprule
\bkhead & \multicolumn{3}{c}{\textbf{QA Task Group}} & \multicolumn{3}{c}{\textbf{Performance Ratio}} \\
\arrayrulecolor{bkInk}\cmidrule(lr){2-4}\cmidrule(lr){5-7}\arrayrulecolor{bkRule}
\bkhead \textbf{Model}
& \textcolor{bkAct!85!black}{\textbf{Coord.\ QA}}
& \textcolor{bkMuted!85!black}{\textbf{MCQ}}
& \textcolor{bkReas!85!black}{\textbf{FIB QA}}
& \textbf{Coord./MCQ}
& \textbf{FIB/MCQ}
& \textbf{Coord./FIB} \\
\midrule
\bkdot{bkCatChem}\enspace ChemVLM-8B~\cite{li2025chemvlm} & \cellcolor{bkAct!5}11.6 & \cellcolor{bkMuted!15}34.9 & \cellcolor{bkReas!10}24.0 & \cellcolor{bkGap!29}0.3 & \cellcolor{bkGap!13}0.7 & \cellcolor{bkGap!21}0.5 \\
\bkdot{bkCatGen}\enspace GPT-5.5~\cite{openai2026gpt55} & \cellcolor{bkAct!17}40.0 & \cellcolor{bkMuted!24}58.3 & \cellcolor{bkReas!27}64.7 & \cellcolor{bkGap!13}0.7 & \cellcolor{bkGap!4}1.1 & \cellcolor{bkGap!17}0.6 \\
\bkdot{bkCatGen}\enspace Claude Opus 4.7~\cite{anthropic2026opus47} & \cellcolor{bkAct!18}41.7 & \cellcolor{bkMuted!26}62.8 & \cellcolor{bkReas!26}62.2 & \cellcolor{bkGap!13}0.7 & 1.0 & \cellcolor{bkGap!13}0.7 \\
\bkdot{bkCatGen}\enspace Qwen3.5-397B-A17B~\cite{qwen35_397b} & \cellcolor{bkAct!9}20.6 & \cellcolor{bkMuted!26}62.2 & \cellcolor{bkReas!27}64.7 & \cellcolor{bkGap!29}0.3 & 1.0 & \cellcolor{bkGap!29}0.3 \\
\bkdot{bkCatEmb}\enspace RoboBrain2.5-8B~\cite{robobrain2026robobrain25} & \cellcolor{bkAct!7}16.6 & \cellcolor{bkMuted!22}52.2 & \cellcolor{bkReas!20}47.5 & \cellcolor{bkGap!29}0.3 & \cellcolor{bkGap!4}0.9 & \cellcolor{bkGap!29}0.3 \\
\bkbottomrule
\end{tabularx}
\end{table}

%% file: figures/Self-explanation.tex
\begin{figure*}[t]
    \centering
    \includegraphics[width=\textwidth]{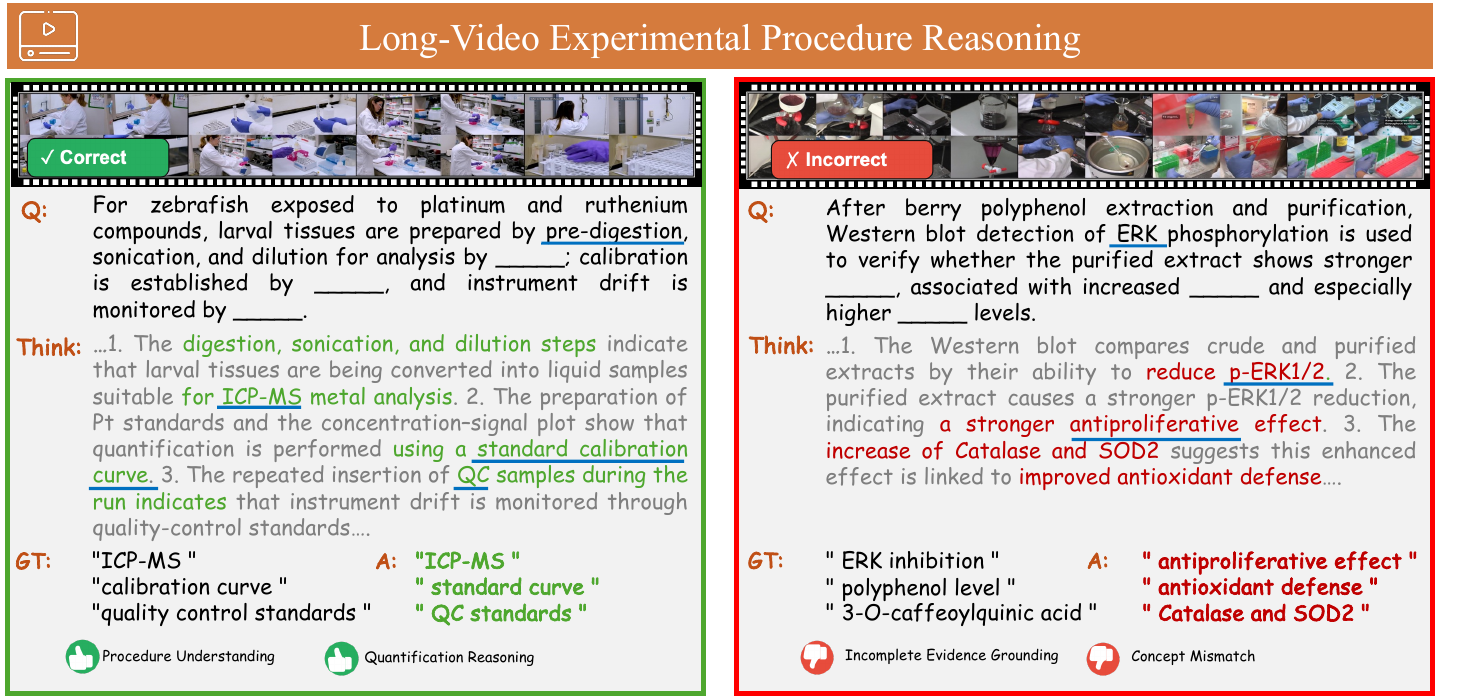}
    \caption{\textbf{Self-explanation examples of GPT-5.4}~\cite{openai2026gpt54} \textbf{on BEAKER.} Representative correct and incorrect cases in Long-Video Experimental Procedure Reasoning (LVPR) show that GPT-5.4 can extract procedural evidence from experimental videos but may struggle to match visual observations with fine-grained chemical concepts. \mbox{GT: Ground Truth.}}
    \label{fig:case_study}
\end{figure*}

%% file: figures/visual_dependence_dumbbell.tex
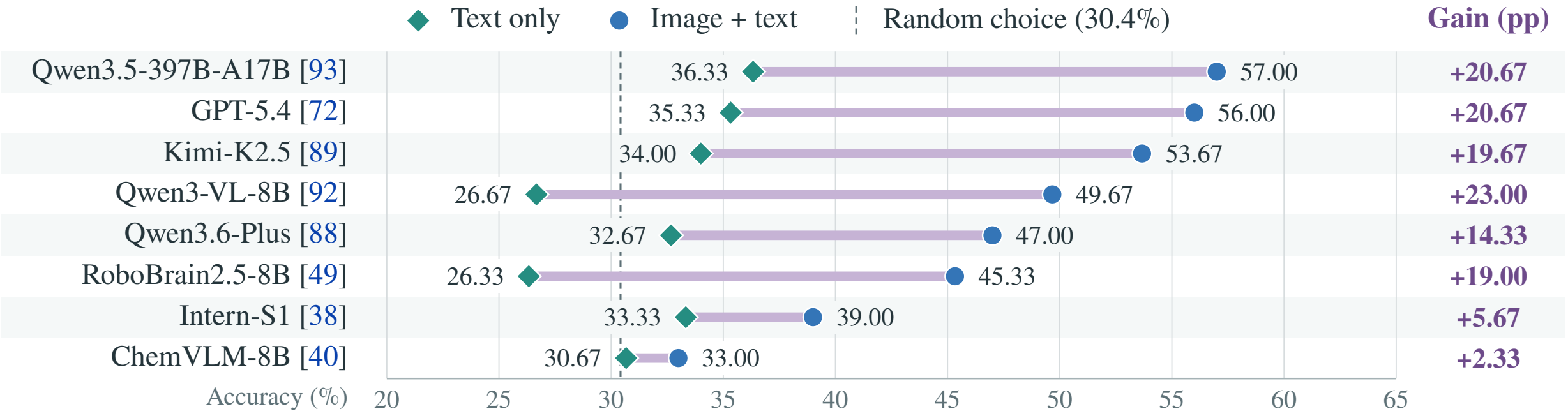
\begin{figure}[htbp]
\centering
\resizebox{\linewidth}{!}{%
\begin{tikzpicture}[x=1pt, y=1pt, font=\fontsize{7.5}{9}\selectfont, text=bkInk]
\def\bkxl{98}\def\bksc{5.7}\def\bkrh{10.4}\def\bkxg{378}
\pgfmathsetmacro{\bkxr}{\bkxl+45*\bksc}
\pgfmathsetmacro{\bkyt}{0.5*\bkrh}
\pgfmathsetmacro{\bkyb}{-7.5*\bkrh}
\pgfmathsetmacro{\bkxrand}{\bkxl+10.42*\bksc}
\def\bkdumbbelldata{%
  0/Qwen3.5-397B-A17B/qwen35_397b/36.33/57.00/+20.67,
  1/GPT-5.4/openai2026gpt54/35.33/56.00/+20.67,
  2/Kimi-K2.5/kimi_k25/34.00/53.67/+19.67,
  3/Qwen3-VL-8B/bai2025qwen3/26.67/49.67/+23.00,
  4/Qwen3.6-Plus/qwen36_2026/32.67/47.00/+14.33,
  5/RoboBrain2.5-8B/robobrain2026robobrain25/26.33/45.33/+19.00,
  6/Intern-S1/bai2025interns1/33.33/39.00/+5.67,
  7/ChemVLM-8B/li2025chemvlm/30.67/33.00/+2.33}
\foreach \i in {0,2,4,6}
  \fill[bkZebra] (0,{-\i*\bkrh-0.5*\bkrh}) rectangle (397.4,{-\i*\bkrh+0.5*\bkrh});
\foreach \k in {20,25,...,65} {
  \pgfmathsetmacro{\bkxk}{\bkxl+(\k-20)*\bksc}
  \draw[bkRule!45, line width=0.3pt] (\bkxk,\bkyt) -- (\bkxk,\bkyb);
  \node[font=\fontsize{6.5}{8}\selectfont, text=bkMuted] at (\bkxk,\bkyb-5.2) {\k};
}
\draw[bkRule, line width=0.4pt] (\bkxl,\bkyb) -- (\bkxr,\bkyb);
\node[anchor=east, font=\fontsize{6.5}{8}\selectfont, text=bkMuted] at (\bkxl-6,\bkyb-5.2) {Accuracy (\%)};
\draw[bkMuted, line width=0.5pt, dash pattern=on 1.6pt off 1.2pt] (\bkxrand,\bkyt) -- (\bkxrand,\bkyb);
\foreach \i/\bknm/\bkkey/\bkt/\bkv/\bkg in \bkdumbbelldata {
  \pgfmathsetmacro{\bky}{-\i*\bkrh}
  \pgfmathsetmacro{\bkxt}{\bkxl+(\bkt-20)*\bksc}
  \pgfmathsetmacro{\bkxv}{\bkxl+(\bkv-20)*\bksc}
  \pgfmathtruncatemacro{\bkodd}{mod(\i,2)}
  \ifnum\bkodd=0 \def\bkbg{bkZebra}\else\def\bkbg{white}\fi
  \node[anchor=east] at (\bkxl-6,\bky) {\bknm~\cite{\bkkey}};
  \draw[bkExp!45, line width=2.4pt] (\bkxt,\bky) -- (\bkxv,\bky);
  \filldraw[fill=bkAct, draw=white, line width=0.4pt] (\bkxt,\bky) +(0,3.1) -- +(3.1,0) -- +(0,-3.1) -- +(-3.1,0) -- cycle;
  \filldraw[fill=bkKnow, draw=white, line width=0.4pt] (\bkxv,\bky) circle (2.5);
  \node[anchor=east, font=\fontsize{6.5}{8}\selectfont, fill=\bkbg, inner xsep=1.4pt, inner ysep=2pt] at (\bkxt-4.6,\bky) {\bkt};
  \node[anchor=west, font=\fontsize{6.5}{8}\selectfont, fill=\bkbg, inner xsep=1.4pt, inner ysep=2pt] at (\bkxv+4.4,\bky) {\bkv};
  \node[font=\fontsize{7}{8}\selectfont\bfseries, text=bkExp!85!black] at (\bkxg,\bky) {\bkg};
}
\def\bkyh{1.25*\bkrh}
\filldraw[fill=bkAct, draw=white, line width=0.4pt] (\bkxl+8,\bkyh) +(0,3.1) -- +(3.1,0) -- +(0,-3.1) -- +(-3.1,0) -- cycle;
\node[anchor=west] (bkl1) at (\bkxl+12.5,\bkyh) {Text only};
\path (bkl1.east) ++(11,0) coordinate (bkc2);
\filldraw[fill=bkKnow, draw=white, line width=0.4pt] (bkc2) circle (2.5);
\node[anchor=west] (bkl2) at ([xshift=4pt]bkc2) {Image + text};
\path (bkl2.east) ++(11,0) coordinate (bkc3);
\draw[bkMuted, line width=0.5pt, dash pattern=on 1.6pt off 1.2pt] ([yshift=-3.6pt]bkc3) -- ([yshift=3.6pt]bkc3);
\node[anchor=west] at ([xshift=3pt]bkc3) {Random choice (30.4\%)};
\node[font=\fontsize{7.5}{9}\selectfont\bfseries, text=bkExp!85!black] at (\bkxg,\bkyh) {Gain (pp)};
\end{tikzpicture}%
}
\caption{%
\textbf{Visual input ablation on image MCQs.}
Accuracy is averaged over five tasks with 60 questions each.
Image + text includes images, questions, and options; Text only
removes the images. Visual input improves average accuracy for
all eight models. The dashed line marks the random-choice baseline
(Appendix~\ref{app:mcq_controls}). Gain denotes the increase in
percentage points (pp).%
}
\label{fig:image_modality_ablation}
\end{figure}

%% file: figures/modality_bar.tex
\begin{figure}[t]
    \centering
    \includegraphics[width=\linewidth]{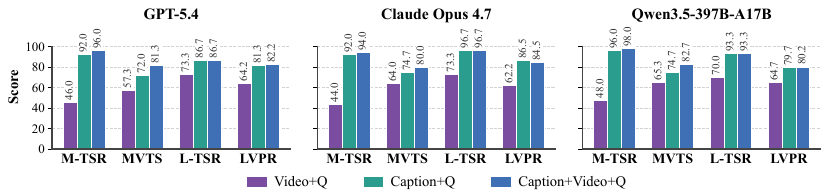}
    \caption{\textbf{Input modality ablation on video-based tasks.}
    Performance of GPT-5.4, Claude Opus 4.7, and Qwen3.5-397B-A17B under video-plus-question, caption-plus-question, and caption-plus-video-plus-question settings. M-TSR: Mid-Video Temporal Sequence Reasoning; MVTS: Mid-Video Task-Step Reasoning; L-TSR: Long-Video Temporal Sequence Reasoning; LVPR: Long-Video Experimental Procedure Reasoning.}
    \label{fig:modality_ablation}
\end{figure}

%% file: table_appendix/Different_frame.tex
\begin{table}[t]
\centering
\caption{\textbf{Frame sampling ablation on video-based BEAKER tasks.}
We evaluate Qwen3.5-397B-A17B~\cite{qwen35_397b} under different frame sampling settings for short-, mid-, and long-video tasks. Frames are listed as short\,/\,mid\,/\,long; long-video frames are arranged into four-frame image grids (16, 32, and 48 grids). The best result in each column is in bold.}
\label{tab:frame_sampling_ablation}
\vspace{1mm}
\small
\setlength{\tabcolsep}{6pt}
\bksetup{1.3}
\colorlet{bkVidS}{bkVid!62}
\colorlet{bkVidM}{bkVid!82}
\begin{tabular}{@{\hspace{5pt}}l c cc cc cc@{\hspace{5pt}}}
\bktoprule
\bkhead & & \multicolumn{2}{c}{\textcolor{bkVidS!70!black}{\textbf{Short video}}} & \multicolumn{2}{c}{\textcolor{bkVidM!78!black}{\textbf{Mid-length video}}} & \multicolumn{2}{c}{\textcolor{bkVid!85!black}{\textbf{Long video}}} \\
\arrayrulecolor{bkVidS}\cmidrule(lr){3-4}\arrayrulecolor{bkVidM}\cmidrule(lr){5-6}\arrayrulecolor{bkVid}\cmidrule(lr){7-8}\arrayrulecolor{bkRule}
\bkhead \textbf{Setting} & \textbf{Frames} & S-TSR$\uparrow$ & SVOR$\uparrow$ & M-TSR$\uparrow$ & MVTS$\uparrow$ & L-TSR$\uparrow$ & LVPR$\uparrow$ \\
\midrule
Fewer Frames & \bkmuted{4\,/\,16\,/\,64} & 82.9 & \textbf{80.8} & \textbf{54.0} & 61.3 & 66.7 & 59.7 \\
\rowcolor{bkVid!7} Standard Frames \bkpill[bkVid]{main} & \bkmuted{8\,/\,32\,/\,128} & 85.7 & 79.2 & 48.0 & \textbf{65.3} & 70.0 & \textbf{64.7} \\
More Frames & \bkmuted{16\,/\,48\,/\,192} & \textbf{88.6} & \textbf{80.8} & 52.0 & 60.0 & \textbf{73.3} & 55.3 \\
\bkbottomrule
\end{tabular}
\end{table}

%% file: sections/5_failure_attribution.tex
\section{Failure Attribution}
\label{sec:failure_attribution}

This section consolidates evidence from the preceding experiments and the appendices to explain why current MLLMs fall short on BEAKER. We first examine why specialized models do not outperform general-purpose ones and then summarize four failure mechanisms shared across model families.

\paragraph{Specialized vs.\ General Models.}
Table~\ref{tab:specialization_gains_losses} compares a chemistry-specialized model (ChemVLM-8B), a science-oriented model (Intern-S1-Mini), and an embodied model (RoboBrain2.5-8B) with the size-matched general-purpose Qwen3-VL-8B. Specialization yields only narrow, task-specific gains: RoboBrain2.5-8B leads only on TRAJ (+4.08) and TSR (+1.11), Intern-S1-Mini leads only on LVPR (+2.50), and each specialized model trails Qwen3-VL-8B on every other task, with the largest gaps for ChemVLM-8B on SVOR ($-$39.20) and SR ($-$35.00). Four factors may jointly explain this pattern, as detailed below.
\begin{enumerate}[label=(\roman*), leftmargin=*]
\item \emph{Domain mismatch.} Chemistry-specialized MLLMs are trained mainly on molecular structures, reaction schemes, and chemical text and figures, and embodied models mainly on general robot-manipulation scenes. Both are far from the real wet-lab imagery, apparatus, and step-by-step procedural videos on which BEAKER is built, so their domain priors may cover little of what it~probes.
\item \emph{Different base models and training recipes.} ChemVLM-8B and Intern-S1-Mini are built on backbones other than Qwen3-VL-8B, so for these two models the size match controls for parameter count but not for base capability. Their lower scores should therefore not be read as evidence that domain fine-tuning by itself erodes general ability; isolating such a specialization--generalization trade-off would require comparing each model with its own base model. RoboBrain2.5-8B instead adopts the Qwen3-VL architecture~\cite{robobrain2026robobrain25}, which makes its comparison with Qwen3-VL-8B closer to a base-versus-specialized contrast; its gains on TRAJ and TSR alongside small losses on the other nine tasks ($-$0.80 to $-$8.33) are consistent with a mild specialization trade-off.
\item \emph{Weakly supervised output spaces.} AFF and TRAJ require open-ended coordinate and trajectory outputs with fine spatial grounding, which domain corpora rarely supervise. Even RoboBrain2.5-8B, whose training targets robotic manipulation, trails the general anchor on AFF ($-$1.81).
\item \emph{Gains track domain overlap.} The few tasks on which a specialized model surpasses Qwen3-VL-8B are those closest to its training domain: trajectory planning and temporal ordering for the embodied RoboBrain2.5-8B, and long-video procedure reasoning, which requires recalling precise scientific terms, for the science-oriented Intern-S1-Mini. The advantage disappears once the task moves away from that domain, which is consistent with a gap driven mainly by domain mismatch rather than by model scale.
\end{enumerate}

\paragraph{Failure Mechanisms Across Models.}
Beyond specialization, four failure mechanisms recur across model families.
\begin{enumerate}[leftmargin=*]
\item \emph{Recognition--grounding gap.} Models handle recognition-style questions considerably better than spatial grounding. Averaged over the 28 models, AFF reaches only 9.7 and TRAJ 34.7 (Table~\ref{tab:benchmark_consolidated_results}), and Actionability is the weakest capability for 27 of the 28 models, averaging 22.2 vs.\ 41.6 for Knowledge (Table~\ref{tab:capability_all}). These tasks require localizing small operable parts, such as valves, nozzles, handles, and connection interfaces, and generating coordinates in an open output space rather than selecting among options; accordingly, all five models in Table~\ref{tab:question_format_analysis} score far lower on Coord.\ QA than on MCQs (Coord./MCQ $\leq$ 0.7).
\item \emph{Insufficient temporal and procedural integration.} M-TSR yields the lowest TSR subset score for 24 of the 28 models (Table~\ref{tab:tsr_length_breakdown}), and expert captions raise all three models in Fig.~\ref{fig:modality_ablation} from 44--48 to 92--96 on M-TSR. Current models therefore struggle less with ordering described steps than with extracting fine-grained operations and step boundaries directly from multi-frame inputs.
\item \emph{Evidence-level mismatch.} In the self-explanation case study (Fig.~\ref{fig:case_study}), GPT-5.4 captures the coarse procedural evidence in the video but answers at the wrong level of granularity, replacing the required component-level answers with broader mechanistic descriptions. Such errors arise from mapping observations to the wrong level of experimental evidence rather than from failing to follow the video.
\item \emph{Lack of chemistry-specific execution grounding.} The training data of general-purpose and embodied models likely cover few of the apparatus, operations, and states specific to chemical laboratories. Consistently, knowledge-oriented image MCQs (ERI, ESS, and SR) average only 40.0--43.9 across the 28 models, against a 30.4\% random-choice baseline and at least 93.3 for human experts, indicating that general capabilities do not transfer directly to execution-oriented, fine-grained laboratory understanding.
\end{enumerate}

Together, these mechanisms indicate that the main bottleneck lies less in recognizing laboratory scenes than in grounding them into operable actions, procedural steps, and chemistry-specific execution, which is precisely the gap BEAKER is designed to expose.

%% file: sections/6_limitations.tex
\section{Limitations and Future Work}
\label{sec:limitations}

While BEAKER provides a systematic evaluation of multimodal models for SDCL execution, several limitations should be noted.

\paragraph{Offline, Single-Turn Scope.}
BEAKER is an offline benchmark that evaluates visual understanding and reasoning from pre-collected images and videos; it does not involve closed-loop interaction with real automated devices, and thus cannot fully capture the adaptive decision-making required in real-time execution. Our evaluation also focuses on single-turn QA and does not assess multi-turn interactive reasoning or error recovery, which are critical for long-horizon autonomous execution. In particular, point and box localization and discrete trajectories are offline approximations of action feasibility, and we do not report the correlation between BEAKER scores and real-robot execution success; BEAKER scores should therefore not be read as evidence of real SDCL deployment capability. Within this scope, offline evaluation separates perception and reasoning failures from controller and hardware errors, keeps evaluation conditions controlled and comparable across models, and avoids executing chemical operations with unvalidated models, which allows BEAKER to localize gaps in the fundamental capabilities on which closed-loop execution depends.

\paragraph{Coverage and Viewpoints.}
Although BEAKER covers 11 task types across four core capability dimensions, its 1,000 QA samples cannot exhaustively represent all chemical experiment scenarios, particularly highly specialized procedures in domains such as electrochemistry or high-pressure reactions. In addition, the visual data are primarily collected from publicly available sources with varying camera perspectives, which may not perfectly align with the egocentric views of specific robotic platforms deployed in real SDCLs.

\paragraph{Potential Source-Level Contamination.}
Because the visual assets are drawn from public datasets and JoVE videos, we cannot fully exclude that some source images or videos appeared in the pretraining data of the evaluated MLLMs. The public technical reports of representative open-source models, such as Qwen3-VL~\cite{bai2025qwen3}, RynnBrain~\cite{dang2026rynnbrain}, and RoboBrain~2.5~\cite{robobrain2026robobrain25}, do not explicitly list JoVE or the other chemical sources used by BEAKER, whereas the training data of closed-source models are undisclosed; a deterministic source-level audit is therefore not possible. However, the source media provide only raw images and videos. All questions, answer options, reference answers, operable regions, point and box annotations, and trajectories in BEAKER were newly created by experts and do not exist in the original sources, so exposure to a source image or video does not imply exposure to a BEAKER question or its answer. Consistent with this, the text-only accuracies of all eight models in Fig.~\ref{fig:image_modality_ablation} remain within 6 points of the 30.4\% random-choice baseline (Appendix~\ref{app:mcq_controls}), so the answers are largely not recoverable from the question text alone. Moreover, even if models have seen some source scenes, they still fail on operable regions, experimental constraints, and execution-level reasoning, which is precisely the gap BEAKER is designed to measure.

\paragraph{Composite Score.}
The Overall score maps heterogeneous metrics, namely MCQ accuracy, IoU for AFF, the DFD-based trajectory score for TRAJ, and expert-judged correctness for LVPR, onto a common 0--100 scale and combines them with fixed capability weights (Section~\ref{sec:benchmarking_procedure}). It is a robustness-checked summary (Appendix~\ref{app:ranking_robustness}) rather than a claim that these metrics are semantically equivalent; task- and capability-level scores (Tables~\ref{tab:benchmark_consolidated_results}, \ref{tab:capability_profile}, and~\ref{tab:capability_all}) should serve as the primary basis for diagnosing model capabilities.

\paragraph{Toward Closed-Loop Evaluation.}
Inspired by the closed-loop interaction paradigm of ESI-Bench~\cite{hong2026esi}, we plan to couple BEAKER with chemical-experiment simulation environments, building on the laboratory-scene and chemical-interaction simulation of LabUtopia~\cite{li2025labutopia} and Chemistry3D~\cite{li2024chemistry3d}, and to extend it into a closed-loop agentic benchmark for SDCLs. In that setting, affordances and trajectories will be executed in simulation rather than scored only by offline coordinates and trajectory distance, the environment will judge whether each action is executable and whether the task is completed, and models will replan from feedback. This will allow BEAKER scores to be validated directly against execution success and will extend the evaluation to decision-making and error correction during dynamic experiments.

%% file: sections/7_conclusion.tex
\section{Conclusion}
\label{sec:conclusion}

We present BEAKER, an expert-curated multimodal benchmark that evaluates, offline and in a single turn, the execution-grounded perception and reasoning of MLLMs as embodied brains in SDCLs. BEAKER decomposes SDCL execution into four capability dimensions---Actionability, Knowledge, Experimentation, and Reasoning---and constructs 1,000 expert-authored QA samples across 11 task types covering both image and video modalities, without AI-assisted sample selection, question generation, or annotation. Our systematic evaluation of 28 MLLMs reveals that current models can follow coarse experimental procedures but remain limited in AFF, TRAJ, fine-grained recognition of laboratory apparatus and states, and mid-length procedural reasoning, exposing a clear capability gap between scene understanding and autonomous decision-making for execution. We plan to extend BEAKER to closed-loop evaluation in chemical-experiment simulation environments. Limitations and future work are discussed in Section~\ref{sec:limitations}, and broader impacts and responsible use in Appendices~\ref{app:broader} and~\ref{app:responsible}.

%% file: sections/appendix.tex
\section{Data Sources}
\label{app:data_sources}

Table~\ref{tab:data_sources} lists all 14 data sources used to construct the visual data in BEAKER, including 12 image sources and 2 video sources, along with their content descriptions and raw sample counts before expert screening. License and usage information for these sources is summarized in Appendix~\ref{app:licenses}.

\input{table_appendix/data_sources}

\section{License and Usage Information for External Assets}
\label{app:licenses}

All image and video data used in BEAKER are collected from publicly accessible sources. Table~\ref{tab:licenses} summarizes the license and usage information for the external assets used in benchmark construction. All image sources specify explicit licenses, with LabPics images, including those re-released in HeinSight 3.0, shared for academic use only, whereas JoVE videos are subject to the JoVE Terms of Use.

\begin{table}[htbp]
\centering
\caption{License and usage information for external assets used in BEAKER construction. Badge colors mark \textcolor{bkAct}{permissive}, \textcolor{bkReas}{use-restricted}, and \textcolor{bkExp}{copyleft} terms.}
\label{tab:licenses}
\small
\setlength{\tabcolsep}{5pt}
\bksetup{1.22}
\begin{tabularx}{\linewidth}{@{\hspace{5pt}}l X l@{\hspace{5pt}}}
\bktoprule
\bkhead \textbf{Source} & \textbf{License / Terms} & \textbf{Platform} \\
\midrule
\bkgroup{bkImg}{3}{Image sources} \\
HeinSight 2.0 \& 3.0~\cite{el2024keeping,el2024eyes} & \bkpill[bkAct]{CC BY 4.0} & \bkmuted{Zenodo} \\
HeinSight 4.0~\cite{ren2026context} & \bkpill[bkAct]{CC BY 4.0} & \bkmuted{Zenodo} \\
ChemEq25~\cite{hossain2025real} & \bkpill[bkAct]{CC BY 4.0} & \bkmuted{Mendeley} \\
Chemical Apparatus~\cite{sasaki2024application} & \bkpill[bkAct]{CC BY 4.0} & \bkmuted{Mendeley} \\
CLAD~\cite{zou2024benchmark} & \bkpill[bkAct]{CC BY 4.0} & \bkmuted{Zenodo} \\
LabPics~\cite{eppel2020labpics} & \bkpill[bkReas]{MIT (academic use only)} & \bkmuted{Zenodo} \\
DenseSSD~\cite{tiong2024machine} & \bkpill[bkAct]{CC BY 4.0} & \bkmuted{Zenodo} \\
DTLD~\cite{wang2024towards} & \bkpill[bkAct]{MIT} & \bkmuted{Hugging Face} \\
Glassware~\cite{sharma2021glassware} & \bkpill[bkAct]{CC BY 4.0} & \bkmuted{Zenodo} \\
Titration~\cite{li2025high} & \bkpill[bkAct]{CC BY 4.0} & \bkmuted{Zenodo} \\
CCSS~\cite{ran2022chemical} & \bkpill[bkAct]{CC BY 4.0} & \bkmuted{Zenodo} \\
Roboflow (10 sets)~\cite{roboflow2024universe} & \bkpill[bkAct]{CC BY 4.0 (8)}\,\bkpill[bkAct]{MIT (1)}\,\bkpill[bkAct]{Public Domain (1)} & \bkmuted{Roboflow} \\
\midrule
\bkgroup{bkVid}{3}{Video sources} \\
IBM Lab Actions~\cite{gabrieli2025activity} & \bkpill[bkAct]{CDLA-Permissive-2.0} & \bkmuted{Zenodo} \\
JoVE Chemistry~\cite{jove_chemistry} & \bkpill[bkReas]{JoVE Terms of Use} & \bkmuted{JoVE} \\
\midrule
\bkgroup{bkMuted}{3}{Tools} \\
FFmpeg~\cite{ffmpeg2024} & \bkpill[bkExp]{LGPL 2.1+ / GPL} & \bkmuted{ffmpeg.org} \\
Whisper~\cite{radford2023whisper} & \bkpill[bkAct]{MIT} & \bkmuted{GitHub} \\
CVAT~\cite{cvat2024} & \bkpill[bkAct]{MIT} & \bkmuted{GitHub} \\
\bkbottomrule
\end{tabularx}
\end{table}

\section{Task-Type Allocation Matrix}
\label{app:allocation_matrix}

The relative proportions of the 11 task types are set by three considerations: (i)~the amount of high-quality real-world visual data that can be collected for each task, (ii)~the importance of the corresponding capability in real SDCL execution, and (iii)~whether experts can annotate the task reliably, together with its annotation difficulty. The final distribution additionally reflects cross-group blind review and consensus adjudication, which retain a different fraction of candidate questions for each task type (Table~\ref{tab:allocation}).

To ensure systematic coverage across all task categories while minimizing annotation redundancy, we design a task-type allocation matrix prior to the question authoring phase.
The matrix prescribes the exact number of questions each of the 10 annotation groups must contribute per task category, so that (i)~the union of all group contributions matches the target distribution across 11 task types, and (ii)~no single group is overburdened with a particular category.

Table~\ref{tab:allocation} presents the complete allocation matrix together with post-review retention statistics.
Each group consists of one chemistry expert and one embodied expert (Section~\ref{sec:annotation}). All groups receive the full curated data pool of 1,617 items and independently select items within their assigned quotas.
For two categories whose totals are not evenly divisible by~10 (M-TSR and MVTS, each totaling~75), five groups are assigned~8 questions and the remaining five are assigned~7, as indicated by the $\dagger$ symbol.
The two senior experts who coordinate overall quality control verify that each group's submission complies with the prescribed allocation before cross-group review begins.

\begin{table}[htbp]
\centering
\caption{%
  Task-type allocation matrix and post-review retention statistics.
  \textbf{Alloc./Grp.}: number of questions assigned to each of the 10 annotation groups.
  \textbf{Initial}: total candidate questions before cross-group review ($= \text{Alloc./Grp.} \times 10$).
  \textbf{Final}: questions retained after cross-group blind review and consensus adjudication.
  \textbf{Ret.}: retention rate.
  \textsc{(s)}/\textsc{(c)}: questions on simple/complex images.
  $\dagger$\,Five groups contribute 8 questions; the other five contribute 7.
  Dots and bars follow the capability colors: \bkdot{bkAct}\,Actionability, \bkdot{bkKnow}\,Knowledge, \bkdot{bkExp}\,Experimentation, \bkdot{bkReas}\,Reasoning.%
}
\label{tab:allocation}
\vspace{4pt}
\small
\setlength{\tabcolsep}{3.4pt}
\bksetup{1.16}
\begin{tabular}{@{}>{\centering\arraybackslash}p{12pt}@{\hspace{5pt}} l ccc r@{\hspace{5pt}}l@{\hspace{3pt}}}
\bktoprule
\bkhead & \textbf{Task Category} & \textbf{Alloc./Grp.} & \textbf{Initial} & \textbf{Final} & \multicolumn{2}{l}{\textbf{Ret. (\%)}} \\
\midrule
\cellcolor{bkHead} & \multicolumn{6}{l}{\bkmuted{\emph{Multiple-Choice Questions (MCQs)}}} \\
\cellcolor{bkHead} & \bkdot{bkKnow}\enspace Equipment and Reagent Identification (ERI) \bkmuted{\textsc{(s)}}   & 4  & 40  & 20  & 50.0 & \bkbar[0.9cm]{50.0}{100}{bkKnow} \\
\cellcolor{bkHead} & \bkdot{bkKnow}\enspace Equipment and Reagent Identification (ERI) \bkmuted{\textsc{(c)}}   & 6  & 60  & 40  & 66.7 & \bkbar[0.9cm]{66.7}{100}{bkKnow} \\
\cellcolor{bkHead} & \bkdot{bkKnow}\enspace Experimental Safety and Standardization (ESS) \bkmuted{\textsc{(s)}} & 4  & 40  & 20  & 50.0 & \bkbar[0.9cm]{50.0}{100}{bkKnow} \\
\cellcolor{bkHead} & \bkdot{bkKnow}\enspace Experimental Safety and Standardization (ESS) \bkmuted{\textsc{(c)}} & 6  & 60  & 40  & 66.7 & \bkbar[0.9cm]{66.7}{100}{bkKnow} \\
\cellcolor{bkHead} & \bkdot{bkKnow}\enspace State Recognition (SR) \bkmuted{\textsc{(s)}}                       & 4  & 40  & 20  & 50.0 & \bkbar[0.9cm]{50.0}{100}{bkKnow} \\
\cellcolor{bkHead} & \bkdot{bkKnow}\enspace State Recognition (SR) \bkmuted{\textsc{(c)}}                       & 6  & 60  & 40  & 66.7 & \bkbar[0.9cm]{66.7}{100}{bkKnow} \\
\cellcolor{bkHead} & \bkdot{bkExp}\enspace Experimental Apparatus Setup (EAS) \bkmuted{\textsc{(s)}}            & 4  & 40  & 20  & 50.0 & \bkbar[0.9cm]{50.0}{100}{bkExp} \\
\cellcolor{bkHead} & \bkdot{bkExp}\enspace Experimental Apparatus Setup (EAS) \bkmuted{\textsc{(c)}}            & 6  & 60  & 40  & 66.7 & \bkbar[0.9cm]{66.7}{100}{bkExp} \\
\cellcolor{bkHead} & \bkdot{bkExp}\enspace Experimental Operations (EOP) \bkmuted{\textsc{(s)}}                 & 4  & 40  & 20  & 50.0 & \bkbar[0.9cm]{50.0}{100}{bkExp} \\
\cellcolor{bkHead} & \bkdot{bkExp}\enspace Experimental Operations (EOP) \bkmuted{\textsc{(c)}}                 & 6  & 60  & 40  & 66.7 & \bkbar[0.9cm]{66.7}{100}{bkExp} \\
\cellcolor{bkHead} & \multicolumn{6}{l}{\bkmuted{\emph{Spatial Annotation (Point, Bounding Box \& Trajectory)}}} \\
\cellcolor{bkHead} & \bkdot{bkAct}\enspace Affordance Grounding (AFF)                                & 30 & 300 & 150 & 50.0 & \bkbar[0.9cm]{50.0}{100}{bkAct} \\
\cellcolor{bkHead}\multirow{-14}{*}{\rotatebox[origin=c]{90}{\textcolor{bkInk}{\textbf{Image (600)}}}}
& \bkdot{bkAct}\enspace Trajectory Planning (TRAJ)                                 & 20 & 200 & 150 & 75.0 & \bkbar[0.9cm]{75.0}{100}{bkAct} \\
\midrule
\cellcolor{bkHead} & \multicolumn{6}{l}{\bkmuted{\emph{Multiple-Choice Questions (MCQs)}}} \\
\cellcolor{bkHead} & \bkdot{bkReas}\enspace Short-Video Temporal Sequence Reasoning (S-TSR)          & 10 & 100 & 70  & 70.0 & \bkbar[0.9cm]{70.0}{100}{bkReas} \\
\cellcolor{bkHead} & \bkdot{bkReas}\enspace Mid-Video Temporal Sequence Reasoning (M-TSR)$^\dagger$  & 7--8 & 75 & 50  & 66.7 & \bkbar[0.9cm]{66.7}{100}{bkReas} \\
\cellcolor{bkHead} & \bkdot{bkReas}\enspace Long-Video Temporal Sequence Reasoning (L-TSR)           & 5  & 50  & 30  & 60.0 & \bkbar[0.9cm]{60.0}{100}{bkReas} \\
\cellcolor{bkHead} & \bkdot{bkExp}\enspace Short-Video Operation Reasoning (SVOR)                    & 15 & 150 & 125 & 83.3 & \bkbar[0.9cm]{83.3}{100}{bkExp} \\
\cellcolor{bkHead} & \bkdot{bkExp}\enspace Mid-Video Task-Step Reasoning (MVTS)$^\dagger$            & 7--8 & 75 & 75  & 100.0 & \bkbar[0.9cm]{100}{100}{bkExp} \\
\cellcolor{bkHead} & \multicolumn{6}{l}{\bkmuted{\emph{Fill-in-the-Blank}}} \\
\cellcolor{bkHead}\multirow{-8}{*}{\rotatebox[origin=c]{90}{\textcolor{bkInk}{\textbf{Video (400)}}}}
& \bkdot{bkReas}\enspace Long-Video Experimental Procedure Reasoning (LVPR)       & 5  & 50  & 50  & 100.0 & \bkbar[0.9cm]{100}{100}{bkReas} \\
\midrule
\bkhead & \textbf{Total} & \textbf{150} & \textbf{1,500} & \textbf{1,000} & \textbf{66.7} & \bkbar[0.9cm]{66.7}{100}{bkInk} \\
\bkbottomrule
\end{tabular}
\end{table}

Several observations emerge from the retention analysis.
First, \emph{simple}-image MCQs exhibit a uniformly lower retention rate (50.0\%) than their \emph{complex} counterparts (66.7\%), reflecting the tendency of simple-scene questions to be answerable through surface-level visual cues---precisely the type flagged for removal during cross-group review (see Section~\ref{sec:annotation}).
Second, the two spatial tasks, whose coordinate-based answers inherently constrain ambiguity, retain 50.0\% (AFF) and 75.0\% (TRAJ) of their candidates; both yield 150 retained questions.
Third, video tasks that require deeper reasoning---MVTS and LVPR (both 100\%) and SVOR (83.3\%)---achieve the highest retention rates, indicating that expert-authored questions grounded in procedural understanding are less susceptible to the quality issues that affect simpler recognition tasks.

\section{Cross-Group Consistency of Spatial Annotations}
\label{app:annotation_agreement}
\label{sec:annotation_agreement}

We compare annotations from the originating group (A) and an independent reviewing group (B) on the same spatial questions. Agreement is measured using box IoU (Eq.~\ref{eq:agree_iou}), bidirectional point-in-region agreement (Eq.~\ref{eq:agree_point}), and trajectory similarity (Eq.~\ref{eq:agree_traj}):
\begin{align}
\operatorname{IoU}(B_A,B_B)
&= \frac{|B_A \cap B_B|}{|B_A \cup B_B|}, \label{eq:agree_iou} \\
\operatorname{PiR}(p_A,p_B)
&= \frac{1}{2}
\left[
\mathbf{1}(p_A \in B_B)
+ \mathbf{1}(p_B \in B_A)
\right], \label{eq:agree_point} \\
S_{\mathrm{traj}}
&= 100 \exp\left(
-\frac{\operatorname{DFD}(P_A,P_B)}{0.40}
\right), \label{eq:agree_traj}
\end{align}
where $B$, $p$, and $P$ denote an annotated box, interaction point, and trajectory, respectively; $\operatorname{PiR}$ denotes Point-in-Region agreement; $\mathbf{1}(\cdot)$ is the indicator function. $\operatorname{DFD}$ denotes the Discrete Fr\'echet Distance~\cite{eiter1994frechet}, computed using the benchmark's trajectory normalization and resampling procedure (Appendix~\ref{app:evaluation}). All metrics are averaged over questions; higher values indicate greater agreement. Table~\ref{tab:annotation_agreement} reports the results for all candidate and retained questions. These scores are not directly comparable with AFF and TRAJ task scores (e.g., the human baseline in Table~\ref{tab:benchmark_consolidated_results}): agreement compares two independent annotations rather than a response with the ground truth, and Eq.~\ref{eq:agree_traj} averages per-question scores, whereas Eq.~\ref{eq:score_traj} applies the exponential to the mean DFD.

\begin{table}[t]
\centering
\caption{Cross-group annotation agreement on candidate and retained
spatial questions. IoU and point-in-region agreement are percentages;
trajectory agreement is scored on a 0--100 scale.}
\label{tab:annotation_agreement}
\small
\setlength{\tabcolsep}{5pt}
\bksetup{1.3}
\newcommand{\bkagv}[3]{\makebox[2.4em][r]{#1}\hspace{5pt}\bkbar[1cm]{#2}{100}{#3}}
\begin{tabular}{@{\hspace{5pt}}l c c c c c@{\hspace{5pt}}}
\bktoprule
\bkhead & \multicolumn{3}{c}{\textcolor{bkAct!80!black}{\textbf{Affordance Grounding (AFF)}}}
& \multicolumn{2}{c}{\textcolor{bkAct!80!black}{\textbf{Trajectory Planning (TRAJ)}}} \\
\arrayrulecolor{bkAct}\cmidrule(lr){2-4}\cmidrule(lr){5-6}\arrayrulecolor{bkRule}
\bkhead \textbf{Subset} & $N$ & Box IoU & Point-in-region & $N$ & Score \\
\midrule
All candidates & 300 & \bkagv{71.1}{71.1}{bkActLight} & \bkagv{87.5}{87.5}{bkActLight} & 200 & \bkagv{80.0}{80.0}{bkActLight} \\
\rowcolor{bkAct!7} \textbf{Retained set} & \textbf{150} & \bkagv{\textbf{86.9}}{86.9}{bkAct} & \bkagv{\textbf{100.0}}{100}{bkAct} & \textbf{150} & \bkagv{\textbf{85.6}}{85.6}{bkAct} \\
\bkbottomrule
\end{tabular}
\end{table}

The retained questions are a subset of the candidates selected through review and adjudication, so this comparison characterizes annotation consistency before and after filtering rather than a causal effect of review. Trajectory agreement measures annotation similarity, not validated physical executability.

Quantitative agreement is reported only for AFF and TRAJ, which are the most prone to spatial ambiguity. For the remaining nine task types, comprising MCQs and the fill-in-the-blank LVPR task, we do not report a single merged agreement coefficient. These tasks follow a protocol of question design, independent blind answering, and consensus adjudication (Section~\ref{sec:annotation}) rather than parallel labeling by multiple annotators in a shared label space, and they span MCQs and fill-in-the-blank questions over images and videos of different lengths, so merging them into one coefficient could be misleading. Instead, every retained question was independently attempted by two non-originating groups without access to the ground truth, and every question flagged during review was revised or discarded through consensus adjudication.

\section{Evaluation Metrics}
\label{app:evaluation}
\label{sec:evaluation}

BEAKER adopts task-specific metrics to accommodate the heterogeneous answer formats across its 11 task types. MCQ tasks are scored by accuracy. For AFF, the model predicts a 2D point, from which a bounding box is constructed with the same dimensions as the ground-truth annotation; performance is the IoU between the predicted and ground-truth boxes (maximum over acceptable targets), scaled to $[0, 100]$. For TRAJ, the predicted and ground-truth keypoint sequences are normalized by image dimensions, and only the predicted sequence is linearly interpolated to the number of ground-truth keypoints, leaving the ground-truth trajectory unchanged. The DFD between the normalized predicted trajectory $\hat{\mathbf{P}}_i$ and the ground-truth trajectory $\mathbf{P}_i$ is computed for each question $i$ and averaged over all $N$ TRAJ questions, and the mean DFD is converted into the task score (Eq.~\ref{eq:score_traj}):
\begin{equation}
\overline{\mathrm{DFD}}=\frac{1}{N}\sum_{i=1}^{N}\mathrm{DFD}(\hat{\mathbf{P}}_i,\mathbf{P}_i),\qquad
\mathrm{Score}_{\mathrm{traj}}=100\exp\!\left(-\frac{\overline{\mathrm{DFD}}}{\tau_d}\right),\quad \tau_d=0.40 .
\label{eq:score_traj}
\end{equation}
Here, $\tau_d$ is the decay scale that controls how rapidly the score decreases as the trajectory distance increases. For the fill-in-the-blank questions of LVPR, each blank is independently evaluated by chemistry experts through manual semantic matching, accepting synonyms and standard domain terminology while rejecting vague or incorrect responses; the per-question score is the average correctness across all blanks. Across 28 evaluated models, all 1,400 fill-in-the-blank evaluation instances are assessed by six chemistry experts, with the workload evenly distributed.

\section{Evaluation Prompt Templates}
\label{app:prompts}

To ensure fair and reproducible evaluation, every model receives the same human-crafted prompt template within each task type. All prompts enforce structured JavaScript Object Notation (JSON) output, reducing the variability caused by free-form generation. Table~\ref{tab:prompt_overview} summarizes the seven templates, and the exact prompts are listed below. TSR is treated as a single task but evaluated separately on short, mid-length, and long videos, resulting in three evaluation subsets (S-TSR, M-TSR, and L-TSR).

\begin{table}[htbp]
\centering
\caption{Overview of evaluation prompt templates. Badges index Prompts 1--7 below and mark image (\bkdot{bkImg}) or video (\bkdot{bkVid}) input; task names follow the capability colors of Table~\ref{tab:allocation}.}
\label{tab:prompt_overview}
\fontsize{8.5}{10}\selectfont
\setlength{\tabcolsep}{4pt}
\bksetup{1.3}
\begin{tabularx}{\linewidth}{@{\hspace{5pt}}c l l l >{\raggedright\arraybackslash}X@{\hspace{5pt}}}
\bktoprule
\bkhead \textbf{\#} & \textbf{Template} & \textbf{Visual Input} & \textbf{Output Format} & \textbf{Tasks} \\
\midrule
\bknum{bkImg}{1} & Image MCQ & 1 image & \texttt{\{"answer":"A"\}} & \textcolor{bkKnow}{ERI}, \textcolor{bkKnow}{ESS}, \textcolor{bkKnow}{SR}, \textcolor{bkExp}{EAS}, \textcolor{bkExp}{EOP} \\
\bknum{bkImg}{2} & Affordance Grounding & 1 image & \texttt{\{"X":123,"Y":456\}} & \textcolor{bkAct}{AFF} \\
\bknum{bkImg}{3} & Trajectory Planning & 1 image & \texttt{\{"trajectory":[\dots]\}} & \textcolor{bkAct}{TRAJ} \\
\bknum{bkVid}{4} & Short-Video MCQ & 8 frames & \texttt{\{"answer":"A"\}} & \textcolor{bkExp}{SVOR}, \textcolor{bkReas}{S-TSR} \\
\bknum{bkVid}{5} & Mid-Video MCQ & 32 frames & \texttt{\{"answer":"A"\}} & \textcolor{bkExp}{MVTS}, \textcolor{bkReas}{M-TSR} \\
\bknum{bkVid}{6} & Long-Video MCQ & 32 grids $\times$ 4 frames & \texttt{\{"answer":"A"\}} & \textcolor{bkReas}{L-TSR} \\
\bknum{bkVid}{7} & Long-Video Fill-in-the-Blank & 32 grids $\times$ 4 frames & \texttt{\{"answer":["a","b"]\}} & \textcolor{bkReas}{LVPR} \\
\bkbottomrule
\end{tabularx}
\end{table}

\begin{bkprompt}[bkImg]{Prompt 1: Image Multiple-Choice}{ERI, ESS, SR, EAS, EOP}
Answer using only the image and options.\\
No reasoning. No explanation. Output exactly one JSON object: \{"answer":"A"\}.\\
Question: \bkph{question}\\
Options:\\
\bkph{options\_block}
\end{bkprompt}

\begin{bkprompt}[bkImg]{Prompt 2: Affordance Point Grounding}{AFF}
Find the best point for this chemistry-lab image question.\\
Use image coordinates with top-left as (0,0), X rightward, Y downward.\\
Coordinates must be integers within width=\bkph{width}, height=\bkph{height}.\\
No reasoning. No explanation. Output exactly one JSON object: \{"X":123,"Y":456\}.\\
Question: \bkph{question}
\end{bkprompt}

\begin{bkprompt}[bkImg]{Prompt 3: Trajectory Planning}{TRAJ}
Plan a safe trajectory for this chemistry-lab image question.\\
Use image coordinates with top-left as (0,0), X rightward, Y downward.\\
Return 4 to 5 key points, integers within width=\bkph{width}, height=\bkph{height}.\\
No reasoning. No explanation. Output exactly one JSON object:\\
\{"trajectory":[\{"X":100,"Y":200\},\{"X":150,"Y":230\},\\
\{"X":200,"Y":260\},\{"X":250,"Y":290\}]\}.\\
Question: \bkph{question}
\end{bkprompt}

\begin{bkprompt}[bkVid]{Prompt 4: Short-Video Multiple-Choice}{SVOR, S-TSR}
These 8 frames come from one short chemistry-lab video in time order.\\
Choose one option using only the frames.\\
No reasoning. No explanation. Output exactly one JSON object: \{"answer":"A"\}.\\
Question: \bkph{question}\\
Options:\\
A. \bkph{option\_a}\ \ B. \bkph{option\_b}\ \ C. \bkph{option\_c}\ \ D. \bkph{option\_d}
\end{bkprompt}

\begin{bkprompt}[bkVid]{Prompt 5: Mid-Video Multiple-Choice}{MVTS, M-TSR}
These 32 frames come from one chemistry-lab video in time order.\\
Choose one option using only the frames.\\
No reasoning. No explanation. Output exactly one JSON object: \{"answer":"A"\}.\\
Question: \bkph{question}\\
Options:\\
A. \bkph{option\_a}\ \ B. \bkph{option\_b}\ \ C. \bkph{option\_c}\ \ D. \bkph{option\_d}
\end{bkprompt}

\begin{bkprompt}[bkVid]{Prompt 6: Long-Video Multiple-Choice}{L-TSR}
These 32 image grids come from one chemistry-lab long video in time order.\\
Choose one option using only the grids.\\
No reasoning. No explanation. Output exactly one JSON object: \{"answer":"A"\}.\\
Question: \bkph{question}\\
Options:\\
A. \bkph{option\_a}\ \ B. \bkph{option\_b}\ \ C. \bkph{option\_c}\ \ D. \bkph{option\_d}
\end{bkprompt}

\begin{bkprompt}[bkVid]{Prompt 7: Long-Video Fill-in-the-Blank}{LVPR}
These 32 image grids come from one chemistry-lab long video in time order.\\
Answer all blanks in order using only the grids.\\
No reasoning. No explanation. Output exactly one JSON object: \{"answer":["a","b"]\}.\\
Question: \bkph{question}\\
Number of blanks: \bkph{blank\_count}
\end{bkprompt}

\section{Capability-Level Scores of All Models}
\label{app:capability_all}

Table~\ref{tab:capability_profile} in the main text reports capability-level scores for 13 representative models. Table~\ref{tab:capability_all} extends these scores to all 28 evaluated models. Averaged over all 28 models, Actionability is the weakest dimension (22.2), followed by Knowledge (41.6), Reasoning (53.8), and Experimentation (58.0). Actionability is also the lowest-scoring dimension for 27 of the 28 models; the only exception is RynnBrain-CoP-8B, whose markedly higher AFF score lifts its Actionability (33.6) above its Knowledge (32.2).

\input{table_appendix/capability_all}

\section{Robustness of Model Scores and Rankings}
\label{app:ranking_robustness}

We assess ranking sensitivity using four alternative aggregations: equal-column macro, equal-group macro, sample-weighted micro, and rank-based aggregation. Equal-column macro averages 13 score columns, with TSR split into short-, mid-, and long-video subtasks. Equal-group macro equally weights the image MCQ, spatial, and video groups. Within each group, task scores are averaged equally; TSR first averages its three video-length subtasks and is then averaged with SVOR, MVTS, and LVPR in the video group. Sample-weighted micro weights each column by its question count, while rank-based aggregation averages within-column model ranks, assigning average ranks to ties.

Table~\ref{tab:aggregation_sensitivity} compares these alternatives
with the official ranking. Spearman correlations are at least
0.980, and top-five/top-ten overlaps are 5/5 and 10/10,
respectively. Rankings are therefore broadly robust to the
tested schemes, although individual positions can change.

\begin{table}[htbp]
\centering
\caption{%
Ranking sensitivity to score aggregation.
Top-$k$ reports overlap with the official top-$k$ set;
average ranks are used for ties.%
}
\label{tab:aggregation_sensitivity}
\vspace{4pt}
\small
\setlength{\tabcolsep}{5pt}
\bksetup{1.3}
\begin{tabularx}{\linewidth}{@{\hspace{5pt}}l >{\raggedright\arraybackslash}X cccc@{\hspace{5pt}}}
\bktoprule
\bkhead \textbf{Aggregation}
& \textbf{Scheme}
& \textbf{Spearman $\boldsymbol{\rho}$}
& \textbf{Kendall $\boldsymbol{\tau}$}
& \textbf{Top-5}
& \textbf{Top-10} \\
\midrule
Equal-column macro & \bkmuted{13 score columns, equal weights}
& 0.994 & 0.952 & \textcolor{bkAct}{5/5} & \textcolor{bkAct}{10/10} \\
\rowcolor{bkZebra} Equal-group macro & \bkmuted{3 task groups, equal weights}
& 0.993 & 0.947 & \textcolor{bkAct}{5/5} & \textcolor{bkAct}{10/10} \\
Sample-weighted micro & \bkmuted{Weighted by question count}
& 0.996 & 0.963 & \textcolor{bkAct}{5/5} & \textcolor{bkAct}{10/10} \\
\rowcolor{bkZebra} Rank-based & \bkmuted{Mean within-column model rank}
& 0.980 & 0.889 & \textcolor{bkAct}{5/5} & \textcolor{bkAct}{10/10} \\
\bkbottomrule
\end{tabularx}
\end{table}

We also perform 10,000 paired, task-stratified bootstrap resamples (Table~\ref{tab:bootstrap_uncertainty}). Questions are sampled with replacement within each task while preserving its size, using identical sampled indices across models. TSR is treated as a single task containing 150 questions during bootstrap resampling. Overall scores and ranks are recomputed from existing predictions under the official aggregation. The 2.5th and 97.5th percentiles define the 95\% score intervals and rank ranges.

\begin{table}[t]
\centering
\caption{%
Bootstrap score and rank uncertainty.
CI denotes the 95\% Confidence Interval from 10,000 paired,
task-stratified resamples. Lower ranks indicate better performance.
Dot colors mark \bkdot{bkCatGen}\,general-purpose, \bkdot{bkCatSci}\,science-oriented, \bkdot{bkCatChem}\,chemistry-specialized, and \bkdot{bkCatEmb}\,embodied brain models; the strip plots each score (dot) and its 95\% CI (bar).%
}
\label{tab:bootstrap_uncertainty}
\vspace{4pt}
\small
\setlength{\tabcolsep}{4.5pt}
\bksetup{1.10}
\begin{tabular}{@{\hspace{5pt}}l c c c c@{\hspace{5pt}}}
\bktoprule
\bkhead \textbf{Model}
& \textbf{Overall}
& \textbf{95\% CI}
& \bkciaxis
& \textbf{95\% Rank Range} \\
\midrule
\bkdot{bkCatGen}\enspace Claude Opus 4.7~\cite{anthropic2026opus47} & 59.2 & [56.7, 61.8] & \bkci{56.7}{61.8}{59.2}{bkCatGen} & 1--2 \\
\rowcolor{bkZebra} \bkdot{bkCatGen}\enspace GPT-5.4~\cite{openai2026gpt54} & 57.2 & [54.5, 59.8] & \bkci{54.5}{59.8}{57.2}{bkCatGen} & 1--4 \\
\bkdot{bkCatGen}\enspace Gemini 3.1 Pro Preview~\cite{google2026gemini31pro} & 55.5 & [53.0, 57.9] & \bkci{53.0}{57.9}{55.5}{bkCatGen} & 2--6 \\
\rowcolor{bkZebra} \bkdot{bkCatGen}\enspace GPT-5.5~\cite{openai2026gpt55} & 55.4 & [52.8, 58.0] & \bkci{52.8}{58.0}{55.4}{bkCatGen} & 2--6 \\
\bkdot{bkCatGen}\enspace Qwen3.5-397B-A17B~\cite{qwen35_397b} & 54.0 & [51.4, 56.6] & \bkci{51.4}{56.6}{54.0}{bkCatGen} & 3--8 \\
\rowcolor{bkZebra} \bkdot{bkCatGen}\enspace Grok-4.20-Reasoning~\cite{xai2026grok420} & 52.9 & [50.4, 55.5] & \bkci{50.4}{55.5}{52.9}{bkCatGen} & 4--10 \\
\bkdot{bkCatGen}\enspace Kimi-K2.5~\cite{kimi_k25} & 52.2 & [49.6, 54.9] & \bkci{49.6}{54.9}{52.2}{bkCatGen} & 5--10 \\
\rowcolor{bkZebra} \bkdot{bkCatGen}\enspace Claude Sonnet 4.6~\cite{anthropic2026sonnet46} & 52.1 & [49.4, 54.7] & \bkci{49.4}{54.7}{52.1}{bkCatGen} & 5--10 \\
\bkdot{bkCatGen}\enspace Kimi-K2.6~\cite{kimik26_2026} & 52.0 & [49.5, 54.6] & \bkci{49.5}{54.6}{52.0}{bkCatGen} & 5--10 \\
\rowcolor{bkZebra} \bkdot{bkCatGen}\enspace Qwen3.6-Plus~\cite{qwen36_2026} & 50.8 & [48.2, 53.4] & \bkci{48.2}{53.4}{50.8}{bkCatGen} & 7--11 \\
\bkdot{bkCatGen}\enspace MiMo-V2-Omni~\cite{mimov2omni_2026} & 48.8 & [46.2, 51.5] & \bkci{46.2}{51.5}{48.8}{bkCatGen} & 10--13 \\
\rowcolor{bkZebra} \bkdot{bkCatGen}\enspace MiMo-V2.5~\cite{mimov25_2026} & 48.2 & [45.5, 50.9] & \bkci{45.5}{50.9}{48.2}{bkCatGen} & 10--14 \\
\bkdot{bkCatGen}\enspace Qwen3-VL-8B~\cite{bai2025qwen3} & 47.1 & [44.5, 49.6] & \bkci{44.5}{49.6}{47.1}{bkCatGen} & 11--15 \\
\rowcolor{bkZebra} \bkdot{bkCatEmb}\enspace RoboBrain2.5-8B~\cite{robobrain2026robobrain25} & 45.3 & [42.7, 47.9] & \bkci{42.7}{47.9}{45.3}{bkCatEmb} & 13--18 \\
\bkdot{bkCatEmb}\enspace ACE-Brain-0-8B~\cite{gong2026ace} & 44.7 & [42.2, 47.4] & \bkci{42.2}{47.4}{44.7}{bkCatEmb} & 13--18 \\
\rowcolor{bkZebra} \bkdot{bkCatGen}\enspace Grok-4.20-Non-Reasoning~\cite{xai2026grok420} & 44.7 & [42.1, 47.4] & \bkci{42.1}{47.4}{44.7}{bkCatGen} & 13--18 \\
\bkdot{bkCatSci}\enspace Intern-S1-Pro~\cite{zou2026interns1pro} & 44.0 & [41.5, 46.6] & \bkci{41.5}{46.6}{44.0}{bkCatSci} & 14--19 \\
\rowcolor{bkZebra} \bkdot{bkCatEmb}\enspace RynnBrain-CoP-8B~\cite{dang2026rynnbrain} & 42.9 & [40.1, 45.6] & \bkci{40.1}{45.6}{42.9}{bkCatEmb} & 15--21 \\
\bkdot{bkCatGen}\enspace InternVL3.5-241B-A28B~\cite{wang2025internvl35} & 41.8 & [39.1, 44.6] & \bkci{39.1}{44.6}{41.8}{bkCatGen} & 16--22 \\
\rowcolor{bkZebra} \bkdot{bkCatGen}\enspace Gemini 2.5 Pro~\cite{google2026gemini25pro} & 40.8 & [38.1, 43.5] & \bkci{38.1}{43.5}{40.8}{bkCatGen} & 18--23 \\
\bkdot{bkCatSci}\enspace Intern-S1~\cite{bai2025interns1} & 40.5 & [37.9, 43.1] & \bkci{37.9}{43.1}{40.5}{bkCatSci} & 18--23 \\
\rowcolor{bkZebra} \bkdot{bkCatEmb}\enspace MiMo-Embodied-7B~\cite{mimo2025embodied} & 39.3 & [36.5, 42.0] & \bkci{36.5}{42.0}{39.3}{bkCatEmb} & 19--24 \\
\bkdot{bkCatEmb}\enspace RynnBrain-8B~\cite{dang2026rynnbrain} & 39.2 & [36.6, 41.9] & \bkci{36.6}{41.9}{39.2}{bkCatEmb} & 19--24 \\
\rowcolor{bkZebra} \bkdot{bkCatEmb}\enspace RynnBrain-Plan-8B~\cite{dang2026rynnbrain} & 37.4 & [34.7, 40.2] & \bkci{34.7}{40.2}{37.4}{bkCatEmb} & 22--26 \\
\bkdot{bkCatSci}\enspace Intern-S1-Mini~\cite{bai2025interns1} & 35.2 & [32.5, 37.9] & \bkci{32.5}{37.9}{35.2}{bkCatSci} & 24--26 \\
\rowcolor{bkZebra} \bkdot{bkCatEmb}\enspace Pelican-VL-1.0-7B~\cite{zhang2025pelicanvl} & 35.2 & [32.6, 37.7] & \bkci{32.6}{37.7}{35.2}{bkCatEmb} & 24--26 \\
\bkdot{bkCatChem}\enspace ChemVLM-26B~\cite{li2025chemvlm} & 30.1 & [27.5, 32.5] & \bkci{27.5}{32.5}{30.1}{bkCatChem} & 27--28 \\
\rowcolor{bkZebra} \bkdot{bkCatChem}\enspace ChemVLM-8B~\cite{li2025chemvlm} & 28.4 & [25.8, 31.0] & \bkci{25.8}{31.0}{28.4}{bkCatChem} & 27--28 \\
\bkbottomrule
\end{tabular}
\end{table}

\section{Temporal Sequence Reasoning Across Video Lengths}
\label{app:tsr_breakdown}

The aggregated TSR score reported in the main paper (Table~\ref{tab:benchmark_consolidated_results}) combines short-, mid-, and long-video temporal ordering tasks into a single metric. To provide finer-grained diagnostic insight, Table~\ref{tab:tsr_length_breakdown} disaggregates this score by video length, revealing how model performance varies with the temporal horizon of the reasoning task.

Several patterns emerge from the breakdown. First, S-TSR is the easiest setting for most models: 20 of the 28 models achieve their highest subset score here, as 4-second clips contain limited temporal complexity and the ongoing or next operation can often be inferred from a few salient visual cues; the remaining eight models score higher on L-TSR. Second, M-TSR is the most challenging setting, yielding the lowest subset score for 24 of the 28 models. Mid-length clips ($\sim$50\,s) span multiple procedural steps whose boundaries are less visually distinct, requiring models to segment continuous operations into discrete steps and reason about their logical dependencies. Third, L-TSR shows high variance: some models recover performance on long videos by leveraging richer contextual cues and explicit experimental milestones (e.g., apparatus changes, reagent additions), while others degrade further, suggesting that their temporal reasoning does not scale with sequence length.

Among model categories, general-purpose MLLMs dominate the leaderboard, with Gemini 3.1 Pro Preview achieving the highest aggregated TSR of 85.9, followed by Grok-4.20-Reasoning at 83.1. Notably, both models maintain strong performance across all three video lengths, indicating robust temporal reasoning capabilities. Embodied brain models show competitive S-TSR and L-TSR scores---RoboBrain2.5-8B reaches 77.1 on S-TSR and 66.7 on L-TSR---but lag behind on M-TSR, where procedural step segmentation is critical. Chemistry-specialized models (ChemVLM-8B/26B) achieve below-average TSR scores (37.9 and 42.4), suggesting that domain-specific chemical knowledge alone does not confer temporal reasoning advantages without sufficient video understanding capabilities. These results reinforce the finding from our main experiments that ordering steps in mid-length procedures remains a key bottleneck for embodied laboratory automation.

\input{table_appendix/overall_TSR}

\section{Multi-Run Stability Analysis}
\label{app:stability}

To assess the reliability of BEAKER scores, we conduct a multi-run stability analysis by evaluating the same model five times under identical benchmark settings. We select Qwen3.5-397B-A17B as the representative model, as it is among the strongest open-source MLLMs in our main experiments. The five runs consist of the main-table run in Table~\ref{tab:benchmark_consolidated_results} and four additional independent runs, all using the same prompts, sampling temperature (0.7), and frame sampling strategy. Table~\ref{tab:multi_run_stability} reports the five-run mean and standard deviation of each task, together with the main-table~scores.

\input{table_appendix/Multiple_rounds}

The Overall score exhibits low variability ($54.2 \pm 0.3$) across the five evaluation runs, and the main-table Overall score (54.0) lies within one standard deviation of this mean; every main-table task score also lies within two standard deviations of the corresponding five-run mean. At the task level, EOP is identical across all runs, and the standard deviations of the other tasks range from 0.2 (AFF) to 1.7 (TSR). For the five image MCQ tasks with 60 questions each, all standard deviations stay below 1.7 points, the score change caused by a single question. The low Overall standard deviation suggests stable aggregate performance for Qwen3.5-397B-A17B across these five runs.

\section{Sanity Checks for Image MCQs}
\label{app:mcq_controls}

We examine all 300 image MCQs, comprising 56 two-option, 27 three-option, and 217 four-option questions. Two-option questions are retained only when both options are diagnostic, as trivial Yes/No judgments are discarded during review (Section~\ref{sec:annotation}). Uniform random selection yields expected accuracies of 50.00\%, 33.33\%, and 25.00\%, respectively, or 30.42\% overall after weighting by question counts. Always selecting the globally most frequent correct-answer position (B, 92 of 300 questions) yields a majority-option baseline of~30.67\%.

Table~\ref{tab:answer_position} reports the ground-truth answer-position distribution and the prediction distributions of \mbox{GPT-5.4} and ChemVLM-8B. Correct answers are roughly balanced within each option-count group.

\begin{table}[htbp]
\centering
\caption{\textbf{Answer-position distributions on image MCQs.} Percentages are computed within each option-count group. $N$ denotes the number of questions; dashes indicate unavailable options. Shading marks deviation from the uniform rate of each group (\textcolor{bkReas}{orange}: above; \textcolor{bkKnow}{blue}: below).}
\label{tab:answer_position}
\vspace{1mm}
\small
\setlength{\tabcolsep}{7pt}
\bksetup{1.2}
\begin{tabular}{@{\hspace{5pt}}l cc cccc@{\hspace{5pt}}}
\bktoprule
\bkhead \textbf{Source} & \textbf{Options} & \textbf{$N$} & \textbf{A (\%)} & \textbf{B (\%)} & \textbf{C (\%)} & \textbf{D (\%)} \\
\midrule
\multirow{3}{*}{\bkdot{bkInk}\enspace Ground truth} & 2 & 56 & \cellcolor{bkReas!3}51.79 & \cellcolor{bkKnow!3}48.21 & \bkmuted{--} & \bkmuted{--} \\
 & 3 & 27 & 33.33 & \cellcolor{bkReas!6}37.04 & \cellcolor{bkKnow!6}29.63 & \bkmuted{--} \\
 & 4 & 217 & 24.42 & 25.35 & 25.35 & 24.88 \\
\midrule
\multirow{3}{*}{\bkdot{bkCatGen}\enspace GPT-5.4~\cite{openai2026gpt54}} & 2 & 56 & \cellcolor{bkKnow!14}41.07 & \cellcolor{bkReas!14}58.93 & \bkmuted{--} & \bkmuted{--} \\
 & 3 & 27 & \cellcolor{bkKnow!6}29.63 & 33.33 & \cellcolor{bkReas!6}37.04 & \bkmuted{--} \\
 & 4 & 217 & 25.81 & \cellcolor{bkReas!4}27.65 & \cellcolor{bkKnow!4}22.58 & \cellcolor{bkKnow!2}23.96 \\
\midrule
\multirow{3}{*}{\bkdot{bkCatChem}\enspace ChemVLM-8B~\cite{li2025chemvlm}} & 2 & 56 & \cellcolor{bkReas!29}67.86 & \cellcolor{bkKnow!29}32.14 & \bkmuted{--} & \bkmuted{--} \\
 & 3 & 27 & 33.33 & \cellcolor{bkReas!36}55.56 & \cellcolor{bkKnow!36}11.11 & \bkmuted{--} \\
 & 4 & 217 & \cellcolor{bkReas!18}36.41 & \cellcolor{bkKnow!5}21.66 & \cellcolor{bkKnow!5}21.66 & \cellcolor{bkKnow!8}20.28 \\
\bkbottomrule
\end{tabular}
\end{table}

\section{Broader Impacts}
\label{app:broader}

BEAKER is designed to advance the safe and reliable deployment of multimodal models in SDCLs. By systematically revealing capability gaps in action grounding, domain-specific laboratory perception, and procedural reasoning, our benchmark can help the community identify failure modes before models are deployed in real laboratory automation, thereby reducing the risk of unsafe or erroneous experimental operations. On the other hand, we acknowledge that the benchmark involves visual data of real chemical experiments, which could, in principle, expose procedural details of hazardous operations. However, all data are collected from publicly available sources, and the benchmark itself does not introduce new chemical knowledge beyond what is already accessible. We also note that benchmark results should not be over-interpreted as certifying a model's readiness for autonomous~execution.

\section{Responsible Use and Dataset Documentation}
\label{app:responsible}

BEAKER is intended for offline evaluation and failure analysis of multimodal embodied brains in SDCL execution contexts. It should not be used to certify models for autonomous laboratory deployment, replace human supervision, or bypass established laboratory safety protocols. Since the visual data are collected from publicly available sources, users should respect the licenses and terms of the original sources listed in Appendix~\ref{app:data_sources} and Appendix~\ref{app:licenses}. Although BEAKER does not introduce new hazardous protocols beyond publicly available materials, some samples may depict chemical operations that require trained personnel and proper safety infrastructure.

%% file: table_appendix/data_sources.tex
\begin{table}[htbp]
\centering
\caption{Data sources for BEAKER. Roboflow encompasses 10 community-contributed datasets.}
\label{tab:data_sources}
\small
\setlength{\tabcolsep}{6pt}
\bksetup{1.2}
\begin{tabularx}{\linewidth}{@{\hspace{5pt}}l X r@{\hspace{5pt}}}
\bktoprule
\bkhead \textbf{Source} & \textbf{Content} & \textbf{\#Raw} \\
\midrule
\bkgroup{bkImg}{3}{Image sources (12)} \\
HeinSight 2.0 \& 3.0~\cite{el2024keeping,el2024eyes} & Vessel detection, liquid/solid phase classification & 5,131 \\
Roboflow~\cite{roboflow2024universe} & Safety labels, apparatus, spills, operations (10 sets) & 3,178 \\
ChemEq25~\cite{hossain2025real} & 25-class real-world apparatus across multiple labs & 1,138 \\
Chemical Apparatus~\cite{sasaki2024application} & Apparatus in complex backgrounds with 6 categories & 629 \\
CLAD~\cite{zou2024benchmark} & 21-class apparatus recognition and detection & 488 \\
LabPics~\cite{eppel2020labpics} & Transparent vessels and material phase segmentation & 281 \\
CCSS~\cite{ran2022chemical} & Chinese chemical safety signs in real-world scenes & 115 \\
DenseSSD~\cite{tiong2024machine} & Transparent vial position detection for lab safety & 61 \\
HeinSight 4.0~\cite{ren2026context} & Context-aware chemical reaction state detection & 48 \\
Glassware~\cite{sharma2021glassware} & Clean-background laboratory glassware classification & 41 \\
DTLD~\cite{wang2024towards} & Transparent liquid level estimation in vessels & 9 \\
Titration~\cite{li2025high} & Automated colorimetric titration monitoring & 9 \\
\rowcolor{bkImg!4} \textbf{Total} & \bkmuted{After screening: 883; final: 410} & \textbf{11,128} \\
\midrule
\bkgroup{bkVid}{3}{Video sources (2)} \\
IBM Lab Actions~\cite{gabrieli2025activity} & Egocentric 4\,s laboratory operation clips (RGB) & 334 \\
JoVE Chemistry~\cite{jove_chemistry} & Long-form experimental procedure recordings & 200 \\
JoVE mid-length segments & Step-level clips segmented from JoVE recordings & 1,753 \\
\rowcolor{bkVid!4} \textbf{Total} & \bkmuted{After screening: 734; final: 400} & \textbf{2,287} \\
\bkbottomrule
\end{tabularx}
\end{table}

%% file: table_appendix/capability_all.tex
\begin{table}[t]
\centering
\caption{\textbf{Capability-level scores of all 28 MLLMs on BEAKER.}
Action., Know., Exper., and Reason.\ denote Actionability, Knowledge, Experimentation, and Reasoning, aggregated from the task scores in Table~\ref{tab:benchmark_consolidated_results} as described in Section~\ref{sec:benchmarking_procedure}; Overall combines them with weights of 0.25, 0.15, 0.35, and 0.25. Cell shading is proportional to the score, and the last row averages each column over all 28 models. Rankings within each column are color-coded: \textcolor{rank1}{\textbf{1st}}, \textcolor{rank2}{\textbf{2nd}}, and \textcolor{rank3}{\textbf{3rd}}.}
\label{tab:capability_all}
\vspace{1mm}
\small
\setlength{\tabcolsep}{5pt}
\bksetup{1.1}
\begin{tabularx}{\linewidth}{@{\hspace{5pt}}X *{4}{>{\centering\arraybackslash}p{1.45cm}} p{0pt} >{\centering\arraybackslash}p{1.65cm}@{\hspace{5pt}}}
\bktoprule
\bkhead & \multicolumn{4}{c}{\textbf{Capability Dimension}} & & \textbf{Aggregate} \\
\arrayrulecolor{bkInk}\cmidrule(lr){2-5}\cmidrule(lr){7-7}\arrayrulecolor{bkRule}
\bkhead \textbf{Model}
& \textcolor{bkAct}{\textbf{Action.}}$\uparrow$
& \textcolor{bkKnow}{\textbf{Know.}}$\uparrow$
& \textcolor{bkExp}{\textbf{Exper.}}$\uparrow$
& \textcolor{bkReas}{\textbf{Reason.}}$\uparrow$
& & \textbf{Overall}$\uparrow$ \\
\midrule
\bkgroup{bkCatChem}{7}{Chemistry-specialized Multimodal Models \textmd{\bkmuted{(open-source)}}} \\
ChemVLM-8B~\cite{li2025chemvlm} & \cellcolor{bkAct!4}11.6 & \cellcolor{bkKnow!11}31.7 & \cellcolor{bkExp!13}37.1 & \cellcolor{bkReas!11}31.0 & & \cellcolor{bkInk!9}28.4 \\
ChemVLM-26B~\cite{li2025chemvlm} & \cellcolor{bkAct!5}15.1 & \cellcolor{bkKnow!10}26.7 & \cellcolor{bkExp!14}38.7 & \cellcolor{bkReas!13}35.2 & & \cellcolor{bkInk!9}30.1 \\
\midrule
\bkgroup{bkCatSci}{7}{Science-oriented Multimodal Models \textmd{\bkmuted{(open-source)}}} \\
Intern-S1~\cite{bai2025interns1} & \cellcolor{bkAct!5}14.0 & \cellcolor{bkKnow!14}37.8 & \cellcolor{bkExp!20}55.1 & \cellcolor{bkReas!17}48.1 & & \cellcolor{bkInk!12}40.5 \\
Intern-S1-Mini~\cite{bai2025interns1} & \cellcolor{bkAct!4}12.0 & \cellcolor{bkKnow!15}41.7 & \cellcolor{bkExp!17}46.8 & \cellcolor{bkReas!14}38.3 & & \cellcolor{bkInk!11}35.2 \\
Intern-S1-Pro~\cite{zou2026interns1pro} & \cellcolor{bkAct!4}10.5 & \cellcolor{bkKnow!16}45.0 & \cellcolor{bkExp!23}63.6 & \cellcolor{bkReas!18}49.6 & & \cellcolor{bkInk!13}44.0 \\
\midrule
\bkgroup{bkCatEmb}{7}{Embodied Brain Models \textmd{\bkmuted{(open-source)}}} \\
ACE-Brain-0-8B~\cite{gong2026ace} & \cellcolor{bkAct!5}12.7 & \cellcolor{bkKnow!17}47.2 & \cellcolor{bkExp!23}64.8 & \cellcolor{bkReas!17}47.2 & & \cellcolor{bkInk!13}44.7 \\
Pelican-VL-1.0-7B~\cite{zhang2025pelicanvl} & \cellcolor{bkAct!6}17.6 & \cellcolor{bkKnow!12}32.8 & \cellcolor{bkExp!19}52.0 & \cellcolor{bkReas!11}30.7 & & \cellcolor{bkInk!11}35.2 \\
RoboBrain2.5-8B~\cite{robobrain2026robobrain25} & \cellcolor{bkAct!6}16.6 & \cellcolor{bkKnow!15}41.7 & \cellcolor{bkExp!23}63.0 & \cellcolor{bkReas!19}51.4 & & \cellcolor{bkInk!14}45.3 \\
RynnBrain-8B~\cite{dang2026rynnbrain} & \cellcolor{bkAct!8}21.9 & \cellcolor{bkKnow!12}33.9 & \cellcolor{bkExp!17}48.1 & \cellcolor{bkReas!17}47.3 & & \cellcolor{bkInk!12}39.2 \\
RynnBrain-CoP-8B~\cite{dang2026rynnbrain} & \cellcolor{bkAct!12}33.6 & \cellcolor{bkKnow!12}32.2 & \cellcolor{bkExp!17}48.4 & \cellcolor{bkReas!18}50.7 & & \cellcolor{bkInk!13}42.9 \\
RynnBrain-Plan-8B~\cite{dang2026rynnbrain} & \cellcolor{bkAct!6}17.5 & \cellcolor{bkKnow!13}35.6 & \cellcolor{bkExp!17}46.2 & \cellcolor{bkReas!17}46.3 & & \cellcolor{bkInk!11}37.4 \\
MiMo-Embodied-7B~\cite{mimo2025embodied} & \cellcolor{bkAct!7}19.4 & \cellcolor{bkKnow!13}35.0 & \cellcolor{bkExp!18}49.9 & \cellcolor{bkReas!17}47.0 & & \cellcolor{bkInk!12}39.3 \\
\midrule
\bkgroup{bkCatGen}{7}{General Multimodal Large Language Models} \\
\multicolumn{7}{@{\hspace{5pt}}l}{\bkmuted{\emph{Closed-source models}}} \\
GPT-5.5~\cite{openai2026gpt55} & \cellcolor{bkAct!14}\textcolor{rank2}{\textbf{40.0}} & \cellcolor{bkKnow!17}48.3 & \cellcolor{bkExp!23}62.6 & \cellcolor{bkReas!23}64.8 & & \cellcolor{bkInk!17}55.4 \\
GPT-5.4~\cite{openai2026gpt54} & \cellcolor{bkAct!13}\textcolor{rank3}{\textbf{36.4}} & \cellcolor{bkKnow!19}\textcolor{rank3}{\textbf{51.7}} & \cellcolor{bkExp!24}67.2 & \cellcolor{bkReas!24}\textcolor{rank3}{\textbf{67.2}} & & \cellcolor{bkInk!17}\textcolor{rank2}{\textbf{57.2}} \\
Claude Opus 4.7~\cite{anthropic2026opus47} & \cellcolor{bkAct!15}\textcolor{rank1}{\textbf{41.7}} & \cellcolor{bkKnow!19}\textcolor{rank1}{\textbf{53.9}} & \cellcolor{bkExp!25}\textcolor{rank3}{\textbf{70.0}} & \cellcolor{bkReas!23}64.7 & & \cellcolor{bkInk!18}\textcolor{rank1}{\textbf{59.2}} \\
Claude Sonnet 4.6~\cite{anthropic2026sonnet46} & \cellcolor{bkAct!12}32.3 & \cellcolor{bkKnow!15}41.7 & \cellcolor{bkExp!23}64.8 & \cellcolor{bkReas!22}60.3 & & \cellcolor{bkInk!16}52.1 \\
Gemini 2.5 Pro~\cite{google2026gemini25pro} & \cellcolor{bkAct!9}25.9 & \cellcolor{bkKnow!14}40.0 & \cellcolor{bkExp!15}40.4 & \cellcolor{bkReas!20}56.7 & & \cellcolor{bkInk!12}40.8 \\
Gemini 3.1 Pro Preview~\cite{google2026gemini31pro} & \cellcolor{bkAct!6}17.2 & \cellcolor{bkKnow!19}\textcolor{rank2}{\textbf{53.3}} & \cellcolor{bkExp!26}\textcolor{rank1}{\textbf{71.6}} & \cellcolor{bkReas!26}\textcolor{rank2}{\textbf{72.6}} & & \cellcolor{bkInk!17}\textcolor{rank3}{\textbf{55.5}} \\
Grok-4.20-Non-Reasoning~\cite{xai2026grok420} & \cellcolor{bkAct!9}23.9 & \cellcolor{bkKnow!12}33.9 & \cellcolor{bkExp!19}54.1 & \cellcolor{bkReas!21}58.8 & & \cellcolor{bkInk!13}44.7 \\
Grok-4.20-Reasoning~\cite{xai2026grok420} & \cellcolor{bkAct!10}26.5 & \cellcolor{bkKnow!16}44.4 & \cellcolor{bkExp!21}59.5 & \cellcolor{bkReas!27}\textcolor{rank1}{\textbf{75.4}} & & \cellcolor{bkInk!16}52.9 \\
MiMo-V2-Omni~\cite{mimov2omni_2026} & \cellcolor{bkAct!9}25.9 & \cellcolor{bkKnow!14}40.0 & \cellcolor{bkExp!23}64.5 & \cellcolor{bkReas!20}55.1 & & \cellcolor{bkInk!15}48.8 \\
Qwen3.6-Plus~\cite{qwen36_2026} & \cellcolor{bkAct!8}23.0 & \cellcolor{bkKnow!14}40.0 & \cellcolor{bkExp!24}66.2 & \cellcolor{bkReas!23}63.4 & & \cellcolor{bkInk!15}50.8 \\
\multicolumn{7}{@{\hspace{5pt}}l}{\bkmuted{\emph{Open-source models}}} \\
Kimi-K2.5~\cite{kimi_k25} & \cellcolor{bkAct!9}24.6 & \cellcolor{bkKnow!18}50.0 & \cellcolor{bkExp!24}67.6 & \cellcolor{bkReas!22}59.8 & & \cellcolor{bkInk!16}52.2 \\
Kimi-K2.6~\cite{kimik26_2026} & \cellcolor{bkAct!9}23.8 & \cellcolor{bkKnow!16}44.4 & \cellcolor{bkExp!24}68.0 & \cellcolor{bkReas!22}62.4 & & \cellcolor{bkInk!16}52.0 \\
MiMo-V2.5~\cite{mimov25_2026} & \cellcolor{bkAct!9}25.7 & \cellcolor{bkKnow!16}43.9 & \cellcolor{bkExp!22}61.7 & \cellcolor{bkReas!20}54.4 & & \cellcolor{bkInk!14}48.2 \\
Qwen3-VL-8B~\cite{bai2025qwen3} & \cellcolor{bkAct!6}15.4 & \cellcolor{bkKnow!17}47.8 & \cellcolor{bkExp!24}65.7 & \cellcolor{bkReas!19}52.2 & & \cellcolor{bkInk!14}47.1 \\
InternVL3.5-241B-A28B~\cite{wang2025internvl35} & \cellcolor{bkAct!6}16.6 & \cellcolor{bkKnow!15}42.2 & \cellcolor{bkExp!19}53.8 & \cellcolor{bkReas!18}50.0 & & \cellcolor{bkInk!13}41.8 \\
Qwen3.5-397B-A17B~\cite{qwen35_397b} & \cellcolor{bkAct!7}20.6 & \cellcolor{bkKnow!18}49.4 & \cellcolor{bkExp!26}\textcolor{rank2}{\textbf{71.1}} & \cellcolor{bkReas!24}66.3 & & \cellcolor{bkInk!16}54.0 \\
\midrule
\bkhead \textbf{Mean (28 models)} & \textbf{22.2} & \textbf{41.6} & \textbf{58.0} & \textbf{53.8} & & \textbf{45.5} \\
\bkbottomrule
\end{tabularx}
\end{table}

%% file: table_appendix/overall_TSR.tex
\begin{table}[t]
\centering
\caption{\textbf{Temporal sequence reasoning across different video lengths.}
S-TSR, M-TSR, and L-TSR denote Temporal Sequence Reasoning on short, mid-length, and long videos, and TSR is their unweighted mean. Cell shading is proportional to the score, and the last row averages each column over all 28 models. Rankings within each column are color-coded: \textcolor{rank1}{\textbf{1st}}, \textcolor{rank2}{\textbf{2nd}}, and \textcolor{rank3}{\textbf{3rd}}. Tied scores are assigned the same rank.}
\label{tab:tsr_length_breakdown}
\vspace{1mm}
\small
\setlength{\tabcolsep}{5pt}
\bksetup{1.1}
\begin{tabularx}{\linewidth}{@{\hspace{5pt}}X *{3}{>{\centering\arraybackslash}p{1.6cm}} p{0pt} >{\centering\arraybackslash}p{1.65cm}@{\hspace{5pt}}}
\bktoprule
\bkhead & \multicolumn{3}{c}{\textcolor{bkReas!80!black}{\textbf{By Video Length}}} & & \textcolor{bkReas!80!black}{\textbf{Aggregate}} \\
\arrayrulecolor{bkReas}\cmidrule(lr){2-4}\cmidrule(lr){6-6}\arrayrulecolor{bkRule}
\bkhead \textbf{Model}
& \textbf{S-TSR}$\uparrow$
& \textbf{M-TSR}$\uparrow$
& \textbf{L-TSR}$\uparrow$
& & \textbf{TSR}$\uparrow$ \\
\midrule
\bkgroup{bkCatChem}{6}{Chemistry-specialized Multimodal Models \textmd{\bkmuted{(open-source)}}} \\
ChemVLM-8B~\cite{li2025chemvlm} & \cellcolor{bkReas!16}45.7 & \cellcolor{bkReas!6}18.0 & \cellcolor{bkReas!18}50.0 & & \cellcolor{bkReas!14}37.9 \\
ChemVLM-26B~\cite{li2025chemvlm} & \cellcolor{bkReas!17}48.6 & \cellcolor{bkReas!8}22.0 & \cellcolor{bkReas!20}56.7 & & \cellcolor{bkReas!15}42.4 \\
\midrule
\bkgroup{bkCatSci}{6}{Science-oriented Multimodal Models \textmd{\bkmuted{(open-source)}}} \\
Intern-S1~\cite{bai2025interns1} & \cellcolor{bkReas!22}61.4 & \cellcolor{bkReas!9}26.0 & \cellcolor{bkReas!13}36.7 & & \cellcolor{bkReas!15}41.4 \\
Intern-S1-Mini~\cite{bai2025interns1} & \cellcolor{bkReas!14}40.0 & \cellcolor{bkReas!8}22.0 & \cellcolor{bkReas!4}10.0 & & \cellcolor{bkReas!9}24.0 \\
Intern-S1-Pro~\cite{zou2026interns1pro} & \cellcolor{bkReas!26}72.9 & \cellcolor{bkReas!6}18.0 & \cellcolor{bkReas!7}20.0 & & \cellcolor{bkReas!13}37.0 \\
\midrule
\bkgroup{bkCatEmb}{6}{Embodied Brain Models \textmd{\bkmuted{(open-source)}}} \\
ACE-Brain-0-8B~\cite{gong2026ace} & \cellcolor{bkReas!24}65.7 & \cellcolor{bkReas!12}34.0 & \cellcolor{bkReas!23}63.3 & & \cellcolor{bkReas!20}54.3 \\
Pelican-VL-1.0-7B~\cite{zhang2025pelicanvl} & \cellcolor{bkReas!16}44.3 & \cellcolor{bkReas!10}28.0 & \cellcolor{bkReas!5}13.3 & & \cellcolor{bkReas!10}28.5 \\
RoboBrain2.5-8B~\cite{robobrain2026robobrain25} & \cellcolor{bkReas!28}77.1 & \cellcolor{bkReas!8}22.0 & \cellcolor{bkReas!24}66.7 & & \cellcolor{bkReas!20}55.3 \\
RynnBrain-8B~\cite{dang2026rynnbrain} & \cellcolor{bkReas!18}50.0 & \cellcolor{bkReas!12}32.0 & \cellcolor{bkReas!22}60.0 & & \cellcolor{bkReas!17}47.3 \\
RynnBrain-CoP-8B~\cite{dang2026rynnbrain} & \cellcolor{bkReas!22}60.0 & \cellcolor{bkReas!14}38.0 & \cellcolor{bkReas!23}63.3 & & \cellcolor{bkReas!19}53.8 \\
RynnBrain-Plan-8B~\cite{dang2026rynnbrain} & \cellcolor{bkReas!19}52.9 & \cellcolor{bkReas!12}32.0 & \cellcolor{bkReas!20}56.7 & & \cellcolor{bkReas!17}47.2 \\
MiMo-Embodied-7B~\cite{mimo2025embodied} & \cellcolor{bkReas!24}65.7 & \cellcolor{bkReas!12}32.0 & \cellcolor{bkReas!14}40.0 & & \cellcolor{bkReas!17}45.9 \\
\midrule
\bkgroup{bkCatGen}{6}{General Multimodal Large Language Models} \\
\multicolumn{6}{@{\hspace{5pt}}l}{\bkmuted{\emph{Closed-source models}}} \\
GPT-5.5~\cite{openai2026gpt55} & \cellcolor{bkReas!30}82.9 & \cellcolor{bkReas!12}32.0 & \cellcolor{bkReas!29}\textcolor{rank2}{\textbf{80.0}} & & \cellcolor{bkReas!23}65.0 \\
GPT-5.4~\cite{openai2026gpt54} & \cellcolor{bkReas!33}\textcolor{rank2}{\textbf{91.4}} & \cellcolor{bkReas!17}46.0 & \cellcolor{bkReas!26}73.3 & & \cellcolor{bkReas!25}\textcolor{rank3}{\textbf{70.3}} \\
Claude Opus 4.7~\cite{anthropic2026opus47} & \cellcolor{bkReas!30}84.3 & \cellcolor{bkReas!16}44.0 & \cellcolor{bkReas!26}73.3 & & \cellcolor{bkReas!24}67.2 \\
Claude Sonnet 4.6~\cite{anthropic2026sonnet46} & \cellcolor{bkReas!26}72.9 & \cellcolor{bkReas!17}\textcolor{rank3}{\textbf{48.0}} & \cellcolor{bkReas!28}\textcolor{rank3}{\textbf{76.7}} & & \cellcolor{bkReas!24}65.8 \\
Gemini 2.5 Pro~\cite{google2026gemini25pro} & \cellcolor{bkReas!23}64.3 & \cellcolor{bkReas!14}38.0 & \cellcolor{bkReas!13}36.7 & & \cellcolor{bkReas!17}46.3 \\
Gemini 3.1 Pro Preview~\cite{google2026gemini31pro} & \cellcolor{bkReas!35}\textcolor{rank1}{\textbf{97.1}} & \cellcolor{bkReas!27}\textcolor{rank2}{\textbf{74.0}} & \cellcolor{bkReas!31}\textcolor{rank1}{\textbf{86.7}} & & \cellcolor{bkReas!31}\textcolor{rank1}{\textbf{85.9}} \\
Grok-4.20-Non-Reasoning~\cite{xai2026grok420} & \cellcolor{bkReas!23}62.9 & \cellcolor{bkReas!17}\textcolor{rank3}{\textbf{48.0}} & \cellcolor{bkReas!29}\textcolor{rank2}{\textbf{80.0}} & & \cellcolor{bkReas!23}63.6 \\
Grok-4.20-Reasoning~\cite{xai2026grok420} & \cellcolor{bkReas!28}78.6 & \cellcolor{bkReas!30}\textcolor{rank1}{\textbf{84.0}} & \cellcolor{bkReas!31}\textcolor{rank1}{\textbf{86.7}} & & \cellcolor{bkReas!30}\textcolor{rank2}{\textbf{83.1}} \\
MiMo-V2-Omni~\cite{mimov2omni_2026} & \cellcolor{bkReas!24}67.1 & \cellcolor{bkReas!15}42.0 & \cellcolor{bkReas!17}46.7 & & \cellcolor{bkReas!19}51.9 \\
Qwen3.6-Plus~\cite{qwen36_2026} & \cellcolor{bkReas!31}\textcolor{rank3}{\textbf{87.1}} & \cellcolor{bkReas!17}\textcolor{rank3}{\textbf{48.0}} & \cellcolor{bkReas!25}70.0 & & \cellcolor{bkReas!25}68.4 \\
\multicolumn{6}{@{\hspace{5pt}}l}{\bkmuted{\emph{Open-source models}}} \\
Kimi-K2.5~\cite{kimi_k25} & \cellcolor{bkReas!29}81.4 & \cellcolor{bkReas!13}36.0 & \cellcolor{bkReas!23}63.3 & & \cellcolor{bkReas!22}60.3 \\
Kimi-K2.6~\cite{kimik26_2026} & \cellcolor{bkReas!29}80.0 & \cellcolor{bkReas!14}40.0 & \cellcolor{bkReas!26}73.3 & & \cellcolor{bkReas!23}64.4 \\
MiMo-V2.5~\cite{mimov25_2026} & \cellcolor{bkReas!28}77.1 & \cellcolor{bkReas!13}36.0 & \cellcolor{bkReas!14}40.0 & & \cellcolor{bkReas!18}51.0 \\
Qwen3-VL-8B~\cite{bai2025qwen3} & \cellcolor{bkReas!28}77.1 & \cellcolor{bkReas!8}22.0 & \cellcolor{bkReas!23}63.3 & & \cellcolor{bkReas!20}54.2 \\
InternVL3.5-241B-A28B~\cite{wang2025internvl35} & \cellcolor{bkReas!19}51.4 & \cellcolor{bkReas!11}30.0 & \cellcolor{bkReas!14}40.0 & & \cellcolor{bkReas!15}40.5 \\
Qwen3.5-397B-A17B~\cite{qwen35_397b} & \cellcolor{bkReas!31}85.7 & \cellcolor{bkReas!17}\textcolor{rank3}{\textbf{48.0}} & \cellcolor{bkReas!25}70.0 & & \cellcolor{bkReas!24}67.9 \\
\midrule
\bkhead \textbf{Mean (28 models)} & \textbf{68.8} & \textbf{37.1} & \textbf{57.0} & & \textbf{54.3} \\
\bkbottomrule
\end{tabularx}
\end{table}

%% file: table_appendix/Multiple_rounds.tex
\begin{table}[htbp]
\centering
\caption{\textbf{Multi-run stability analysis on BEAKER.} 
We report the mean and sample Standard Deviation (SD, computed with an $n-1$ denominator) over five evaluation runs of Qwen3.5-397B-A17B~\cite{qwen35_397b} under the same benchmark setting, comprising the main-table run (Table~\ref{tab:benchmark_consolidated_results}) and four additional independent runs. The main-table run is also listed for~reference. $\Delta$ is the main-table run minus the five-run mean; the strip plots $\Delta$ (dot) against the $\pm$1\,SD (dark) and $\pm$2\,SD (light) bands, in score points.}
\label{tab:multi_run_stability}
\vspace{2mm}
\small
\setlength{\tabcolsep}{5pt}
\bksetup{1.4}
\begin{tabular}{@{\hspace{5pt}}l l c c c r c@{\hspace{5pt}}}
\bktoprule
\bkhead \textbf{Capability} & \textbf{Task} & \bkstack{\textbf{Five-run}}{\textbf{mean}} & \textbf{SD} & \bkstack{\textbf{Main-table}}{\textbf{run}} & $\boldsymbol{\Delta}$ & \bkdevaxis \\
\midrule
\multirow{2}{*}{\textcolor{bkAct}{\textbf{Actionability}}}
& AFF  & 3.9  & 0.2 & 3.8  & $-0.1$ & \bkdev{0.2}{-0.1}{bkAct} \\
& TRAJ & 37.8 & 0.5 & 37.3 & $-0.5$ & \bkdev{0.5}{-0.5}{bkAct} \\
\midrule
\multirow{3}{*}{\textcolor{bkKnow}{\textbf{Knowledge}}}
& ERI  & 51.0 & 0.9 & 50.0 & $-1.0$ & \bkdev{0.9}{-1.0}{bkKnow} \\
& ESS  & 52.3 & 0.9 & 53.3 & $+1.0$ & \bkdev{0.9}{1.0}{bkKnow} \\
& SR   & 43.0 & 1.4 & 45.0 & $+2.0$ & \bkdev{1.4}{2.0}{bkKnow} \\
\midrule
\multirow{4}{*}{\textcolor{bkExp}{\textbf{Experimentation}}}
& EAS  & 66.7 & 1.2 & 65.0 & $-1.7$ & \bkdev{1.2}{-1.7}{bkExp} \\
& EOP  & 71.7 & 0.0 & 71.7 & $0.0$  & \bkdev{0.0}{0.0}{bkExp} \\
& SVOR & 80.0 & 0.8 & 79.2 & $-0.8$ & \bkdev{0.8}{-0.8}{bkExp} \\
& MVTS & 64.5 & 1.2 & 65.3 & $+0.8$ & \bkdev{1.2}{0.8}{bkExp} \\
\midrule
\multirow{2}{*}{\textcolor{bkReas}{\textbf{Reasoning}}}
& TSR  & 69.3 & 1.7 & 67.9 & $-1.4$ & \bkdev{1.7}{-1.4}{bkReas} \\
& LVPR & 64.5 & 1.0 & 64.7 & $+0.2$ & \bkdev{1.0}{0.2}{bkReas} \\
\midrule
\bkhead \multicolumn{2}{@{\hspace{5pt}}l}{\textbf{Overall}} & \textbf{54.2} & \textbf{0.3} & \textbf{54.0} & $-0.2$ & \bkdev{0.3}{-0.2}{bkInk} \\
\bkbottomrule
\end{tabular}
\end{table}